\documentclass[10pt,journal,compsoc]{IEEEtran}

\usepackage{bm}
\usepackage{cite}
\usepackage{algorithmic}
\usepackage{graphicx}
\usepackage{textcomp}
\usepackage{epstopdf}
\usepackage{booktabs}
\usepackage{amsmath}
\usepackage{hyperref}
\usepackage{tipa}
\usepackage{esint}
\usepackage{bigstrut}
\usepackage{arydshln}
\usepackage{multirow}
\usepackage{cite}
\usepackage{amssymb}
\usepackage{array}
\usepackage{color}
\usepackage{colortbl}
\usepackage{cite}
\usepackage{stfloats}
\usepackage{makecell}
\usepackage{soul}
\usepackage{diagbox}
\usepackage[numbers,sort&compress]{natbib}
\usepackage[table,xcdraw]{xcolor}
\hypersetup{hidelinks} 
\usepackage{tablefootnote}
\usepackage{colortbl} 
\usepackage{xcolor} 
\usepackage{array}
\usepackage{verbatim}

\graphicspath{{Figures/}}

\begin{document}
	
\bibliographystyle{unsrt}

\title{Detecting Generated Images by Real Images Only}

 \author{Xiuli~Bi, Bo~Liu,~\IEEEmembership{Member,~IEEE,} Fan~Yang, Bin~Xiao, Weisheng~Li,~\IEEEmembership{Member,~IEEE,} Gao~Huang,~\IEEEmembership{Member,~IEEE,}
Pamela~C.~Cosman,~\IEEEmembership{Fellow,~IEEE} 

        

\IEEEcompsocitemizethanks{\IEEEcompsocthanksitem Xiuli Bi, Bo Liu, Fan Yang, Bin Xiao and Weisheng Li are with the Department of Computer Science and Technology, Chongqing University of Posts and Telecommunications, Chongqing, China. email: bixl@cqupt.edu.cn, boliu@cqupt.edu.cn, cquptfan@gmail.com, xiaobin@cqupt.edu.cn, liws@cqupt.edu.cn.\protect
\IEEEcompsocthanksitem Gao Huang is with the Department of Automation, Tsinghua University, Beijing 100084, China. email: gaohuang@tsinghua.edu.cn
\IEEEcompsocthanksitem Pamela C. Cosman is with the Department of Electrical and Computer Engineering, University of California at San Diego, La Jolla, CA 92093 USA. e-mail: pcosman@ucsd.edu.

\IEEEcompsocthanksitem This work was supported in part by the National Key Research and Development Project under Grant 2019YFE0110800, in part by the National Natural Science Foundation of China under Grant 62172067 and Grant 61976031, the Natural Science Foundation of Chongqing for Distinguished Young Scholars under Grant CSTB2022NSCQ-JQX0001. Corresponding author: Bin Xiao.}
\thanks{Manuscript received April 19, 2005; revised August 26, 2015.}}

%
%

\markboth{Journal of \LaTeX\ Class Files,~Vol.~14, No.~8, August~2015}%
{Shell \MakeLowercase{\textit{et al.}}: Bare Advanced Demo of IEEEtran.cls for IEEE Computer Society Journals}

\IEEEtitleabstractindextext{%
\begin{abstract}
As deep learning technology continues to evolve, the images yielded by generative models are becoming more and more realistic, triggering people to question the authenticity of images. Existing generated image detection methods detect visual artifacts in generated images or learn discriminative features from both real and generated images by massive training. This learning paradigm will result in efficiency and generalization issues, making detection methods always lag behind generation methods. This paper approaches the generated image detection problem from a new perspective: Start from real images. By finding the commonality of real images and mapping them to a dense subspace in feature space, the goal is that generated images, regardless of their generative model, are then projected outside the subspace. As a result, images from different generative models can be detected, solving some long-existing problems in the field. Experimental results show that although our method was trained only by real images and uses 99.9\% less training data than other deep learning-based methods, it can compete with state-of-the-art methods and shows excellent performance in detecting emerging generative models with high inference efficiency. Moreover, the proposed method shows robustness against various post-processing. These advantages allow the method to be used in real-world scenarios.
\end{abstract}

}

\maketitle

\section{Introduction}\label{1}

\begin{figure}[t]
	\centering
	\includegraphics[height=3.9cm]{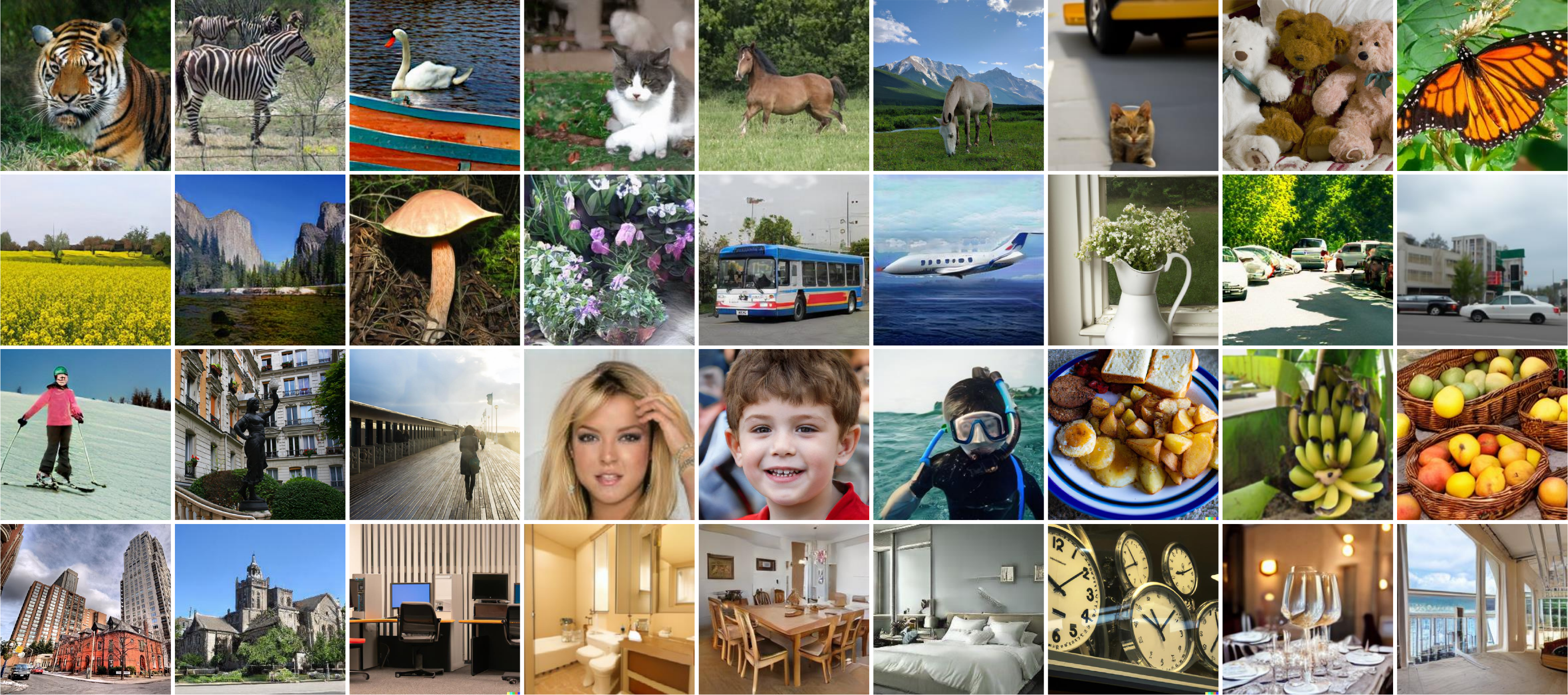}
	\caption{Which are generated images and which are real images captured by cameras?}
	\label{fig:example1}
\end{figure}

\IEEEPARstart{C}{urrently}, the popularity of deep neural networks has driven the rapid development of digital forgery technology, making it easy to abuse AI synthesis algorithms. Various eye-popping technologies have entered our lives, from image content manipulation to scene synthesis, from face attribute tampering to face swapping. In Fig.~\ref{fig:example1}, can you discern which images are captured by cameras and which are generated by neural networks? In fact, all images in Fig.~\ref{fig:example1} are generated by generative models, such as Generative Adversarial Networks (GAN), flow models, and diffusion models. These generated fake images can be used as fun plugins for applications such as face makeup~\cite{liu2021psgan++} and attribute editing~\cite{shen2020interfacegan}, but also to spread falsehoods. For example, unscrupulous people send LinkedIn job postings by impersonating real people with synthetic faces to conduct fraudulent activities. Since many images produced by generative models can reach the point of deceiving human eyes, detection methods that can distinguish real and generated images are strongly needed. In this paper, we use the term ``real images" to mean photographs of the real world as captured by digital cameras based on visible light. This definition excludes images of the real world from other sorts of detectors, such as CT and MR scans.

\begin{figure*}[t]
	\centering
	\includegraphics[width=0.9\textwidth]{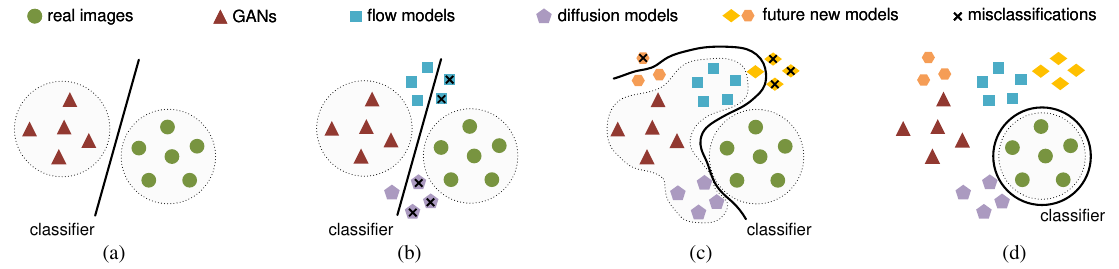}
	\caption{Our method adopts a new perspective based on real images to detect generated images. The dashed circles indicate the samples in the training set. (a) The feature space determined by real and GAN-generated images; (b) The space does not fit for all generative models, such as flow and diffusion model-generated images; (c) The feature space becomes complicated to include more generative models: the classifier will be more complex and fail to handle unseen new models, and the training costs are high; (d) Our method only uses real images to construct a dense subspace and performs one-class classification to detect generated images, and thus is able to detect existing and possible future new generative models.}
	\label{fig:ourmethod}
\end{figure*}

Existing detection methods~\cite{zhang2019detecting,frank2020leveraging,liu2020global,dang2020detection,zhao2021multi,chai2020makes,durall2020watch,wang2020cnn, gragnaniello2021gan, yu2019attributing, dzanic2020fourier} are designed for specific types of GAN-generated images. No matter what strategy these methods take for training, they try to find the decision boundaries between real and specific GAN-generated images, as shown in Fig.~\ref{fig:ourmethod}(a). In the feature space constructed by training balanced real and specific GAN-generated images, the classifier can accurately classify them; these detection methods are very effective in the types of GAN-generated images in their training set. When they encounter images generated by GAN models not seen in the training set, using images generated by ProGAN~\cite{karras2017progressive} for training can avoid a significant performance drop.

However, as shown in Fig.~\ref{fig:ourmethod}(b), the constructed feature space and classifier, which are based on GAN-generated images, will not fit other model-generated images well, such as flow model- and diffusion model-generated images. Experimental results validated this view~\cite{zhang2019detecting,frank2020leveraging,liu2020global,dang2020detection,zhao2021multi,chai2020makes,durall2020watch,wang2020cnn,gragnaniello2021gan,yu2019attributing,dzanic2020fourier}: the classes of the generated images in the training set determine the detection performance. To enhance the versatility and detection performance of the classifier, one might think that adding more types of generated images to construct a proper feature space by re-training or fine-tuning will fit existing popular generative models. However, as shown in Fig.~\ref{fig:ourmethod}(c), the space becomes much more complicated, and such a classifier is required to fit the increasingly complicated feature space, resulting in a more difficult and time-consuming training process. By adopting this strategy, the classifier is chasing future new generative models, reducing the practical utilization value in realistic scenarios; this strategy also triggers more problems which will be discussed in Subsection~\ref{futile}. 

These problems are caused by considering generated image detection from the perspective of generated images. In this paper, we analyze the characteristics of real and generated images, and find that the noise of real images is very similar while that of generated images is different. Based on this, we model the generated image detection problem as a one-class classification problem. By finding the commonality of real images and mapping them to a dense subspace in this noise-feature space, the goal is that generated images, regardless of their generative models, should ideally be mapped outside the subspace and consequently detected as shown in Fig.~\ref{fig:ourmethod}(d).

The contributions of the paper can be summarized as follows:
\begin{itemize}
    \item We are the first to propose a new perspective on generated image detection: Start from real images. In this way, long-existing problems such as poor generalization, and high training and inference costs, can be handled. 
        
    \item We have analyzed an appropriate way to map real images to a dense subspace theoretically and experimentally so that our method achieves good performance while using 99.9\% less training data. It also has good robustness against various post-processing operations. Even for emerging generative models, the detection performance is very good. Thus it can be used in realistic scenarios. 

    \item This paper improves our previous work \textit{Detecting Generated Images by Real Images} published in ECCV'22, which still needs real and generated images for massive training. This paper learns from real images only while keeping superior detection performance and achieves a much better inference efficiency.
\end{itemize}

\par In the remainder of the paper, we first analyze the work related to generated image detection in Section~\ref{2}. Then in Section~\ref{3} we model generated image detection as a one-class classification 
problem and introduce a proper way to map real images to a subspace in feature space based on the commonality of real images. We present and analyze the experimental results in Section~\ref{4}, and draw conclusions in Section~\ref{6}.

\section{Related Work}\label{2}
\subsection{Existing Generated Image Detection Methods}\label{2.1}
Existing generated image detection methods mainly rely on deep learning. They can be divided into image-artifact-detection-based and massive-training-based approaches. The artifact detection approach probes the traces left by the generative networks. Using these traces in the spatial or frequency domain, a trained classifier can discern the generated images. In contrast, by deploying massive training, the second group of approaches takes advantage of the strong learning ability of deep neural networks to search for the decision boundaries between real and generated images. 

\subsubsection{Using Artifacts for Detection}
\par For methods focusing on image artifacts, Yu et al.~\cite{yu2019attributing} argue that each GAN model has a unique fingerprint due to the influence of training data, network structure, loss function, parameter settings, and other factors. Liu et al.~\cite{liu2020global} notice a difference in texture between real and generated faces and used Gram matrices in the network to capture important and long-distance textural information. Dang et al.~\cite{dang2020detection} recognize the importance of the spatial information of the tampered region and use the attention mechanism to highlight informative regions to improve binary classification results. Zhao et al.~\cite{zhao2021multi} consider the generated image detection task as a fine-grained problem and improve detection performance by using the attention mechanism to amplify subtle differences in the shallow layers. It is argued in~\cite{chai2020makes} that extracting forensic features from local areas is more effective than a global approach.  

\par Zhang et al.~\cite{zhang2019detecting} introduce a generator that can simulate sampling artifacts on several commonly seen GANs. This method shows that studying the artifacts in the frequency domain is an effective way to expose generated images. Durall et al.~\cite{durall2020watch} find the distribution of high-frequency components of real images hard to simulate by generative models. Therefore generated images can be classified according to the characteristics of their high-frequency components. Dzanic et al.~\cite{dzanic2020fourier} argue that generated images do not decay at the highest frequencies while real images do, and images could be detected based on the degree of partial decay at high frequencies. Frank et al.~\cite{frank2020leveraging} demonstrate that the artifacts in GAN-generated images are caused by upsampling operations, using DCT transforms to detect them. 

\par The above methods analyze generative models and extract discriminative artifact features for detection. However, as generative models evolve and the quality of generated images improves, such as through the emergence of diffusion models, extracting discriminative artifacts from generated images becomes more difficult. Also, these spatial and frequency domain methods do not generalize well to unseen generative models, resulting in an unsatisfactory performance for cross-dataset experiments.

\subsubsection{Deploying Massive Training for Detection}\label{2.3}
\par The massive training-based methods overcome some shortcomings of the artifact detection methods. By deploying massive training from generated and real images, discriminative features of generated images are automatically identified. In Wang's method~\cite{wang2020cnn}, 720k real images were used with the generated images of ProGAN~\cite{karras2017progressive} to train ResNet50~\cite{he2016deep}, which can be generalized to detect some generative models. Based on this, Gragnaniello et al.~\cite{gragnaniello2021gan} used a modified ResNet50 network by reducing two downsampling layers to improve the detection performance. However, the training cost surges. Due to the use of neural networks to automatically find differences between real and generated images, such methods are easily influenced by the generated images in the training set. Thus, it does not work to detect other generative models such as flow-based models~\cite{kingma2018glow,dinh2014nice,dinh2016density,papamakarios2017masked,kingma2016improved} and diffusion models~\cite{sohl2015deep,song2019generative,ho2020denoising,rombach2021highresolution,rombach2022high}. 

\subsection{One-class Classification Problems}\label{2.2}
\par Typical classifiers like SVM are two- or multi-class classifiers which generalize to new data by maximizing a margin between existing training data from different classes~\cite{boser1992training} in some feature space. One-class classifiers decide a boundary using one class of data. Any data outside the boundary will be classified as a different class. There are two implementations to construct the decision boundary. The first, proposed by Schölkopf et al.~\cite{scholkopf1999support}, maximizes the distance between a hyperplane decided by support vectors and the coordinate origin. The second, by Tax and Duin, restricts training data to a hypersphere~\cite{tax2004support}. One-class classification problems commonly exist in novelty and outlier detection~\cite{Zhai2023etd,zhang2023federated,lo2023adv,zaheer2022stab}. Successful application of one-class classifiers depends on whether the feature space is well constructed. An improper feature space contains large voids between the class instances, resulting in model failure or producing many false classifications.

\section{Using Real Images to Detect Generated Images}\label{3}

\par From Fig.~\ref{fig:ourmethod}, performing generated image detection from the perspective of generated images leads to many problems. In our definition, real images are produced by cameras, as a physical imaging process. In contrast, generated images are generated by deep learning networks, as a computational process. We note that the division into physical versus computational imaging processes is not a clean separation. On the one hand, camera images from physical imaging undergo various computational steps in the camera pipeline such as white balancing, demosaicking from a color filter array, or JPEG compression, and on the other hand, conditional generative models can take a real image as input to a neural network. Therefore, we consider this task from real images by starting with their properties. As shown in Fig.~\ref{fig:ourmethod}(d), from the perspective of real images, the classifier can distinguish various types of generated images. That is, it becomes a one-class classification problem.

\subsection{Modeling the One-class Classification Problem}\label{model_one_class}

\par We consider a sequence $\mathcal{T}$ consisting of some real images $t_1, t_2,..., t_i,..., t_{\ell}$, where $\ell \in \mathbb{N}$ is the number of images. Suppose a mapping $\Phi:\mathcal{T} \rightarrow \mathcal{S}$, where the dot product in the feature of $\Phi$ can be computed by a kernel $K$:
\begin{equation}
\label{dandkernel}
\Phi(t_i) \cdot \Phi(t_j)=K(t_i,t_j)
\end{equation}
where $K$ can be a simple kernel such as a Radial Basis Function (RBF) kernel:
\begin{equation}
\label{gaussiankernel}
    K(t_i,t_j)=e^{-\Vert t_i - t_j \Vert ^2 /2\sigma^2}.
\end{equation}

\par To create a dense subspace $\mathcal{S}$ that contains real images only, we can limit these real images within a hypersphere. Or, more conveniently, we can separate the real images using a hyperplane parameterized by vector normal $\bm{w}$ and offset $\rho$ from the origin by optimizing
\begin{equation}\label{oceq}
\begin{split}
\min_{\bm{w},\bm{\eta},\rho} \frac{1}{2}\Vert \bm{w} \Vert^2+\frac{1}{v \ell}\sum_i \eta_i - \rho, \\
s.t. \text{ }  \bm{w} \cdot \Phi(t_i) \geq \rho - \eta_i, \eta_i \geq 0,
\end{split}
\end{equation}
to maximize the distance $\theta$ between the origin and this hyperplane. Then $\mathcal{S}$ will be the space which satisfies $\bm{w} \cdot \Phi(t_i) \geq \rho$. In Eq.~\ref{oceq}, the column vector $\bm{\eta}  = [{\eta _1},{\eta _2}, \ldots ,{\eta _\ell}]$ and each $\eta _i$ is the slack variable corresponding to the $i$-th real image; $v\in(0,1)$ is a trade-off parameter between optimization terms. Finally, for an image $I\notin \mathcal{T}$, the decision function:
\begin{equation}\label{df}
  \varphi(I)=\mathrm{sgn}\left[\bm{w} \cdot \Phi(I) - \rho\right]
\end{equation}
will decide whether the image $I$ is in $\mathcal{S}$ or not. For real images $t_i$ in the sequence $\mathcal{T}$, the above sign function will be positive. Using the kernel trick in Eq.~\ref{dandkernel}, the sign function in Eq.~\ref{df} can also be denoted in the form of support vector expansion:
\begin{equation}\label{signfunction}
    \varphi(I)=\mathrm{sgn}\left[\sum_i \alpha_i K(I, t_i) - \rho\right].
\end{equation}

\begin{figure}[t]
	\centering
	\includegraphics[height=6.3cm]{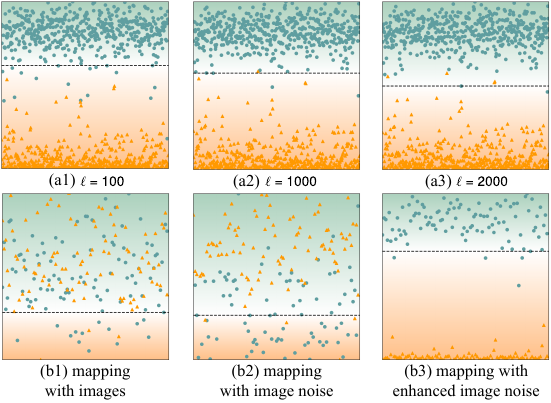}
	\caption{Different values of $\ell$ and mapping functions $f\in\mathcal{F}$ affect the decision boundary and the classifier performance. Green circles and orange triangles represent real and generated images, respectively.}
	\label{mapping_function}
\end{figure}

\par To determine whether real images are mapped to $\mathcal{S}$ via $\Phi$, we analyze the probability $P$ that a real image $t_i\in \mathcal{T}$ falls outside $\mathcal{S}$ after mapping. For any $\theta \in \mathbb{R}$, according to Theorem 7 in~\cite{scholkopf2001estimating} (proved by Lemma 3.9 in~\cite{shawe1998structural}), all images $t_1, t_2,...t_i,..., t_{\ell} \in \mathcal{T}$ are mapped to $\mathcal{S}$ with probability $1-\delta$ if $f(t_i) \geq \theta + \gamma$ holds for any $t_i$
(given that there is a class $\mathcal{F}$ of functions which map $\mathcal{T}$ to $\mathbb{R}$, and $f \in \mathcal{F}$). The mapping methods are not limited to the one we introduced, and we use $f$ as some function in the set. The probability $P$ that the mapping of the real image $t_i$ falls outside $\mathcal{S}$ is 
\begin{equation}\label{firstrisk}
    P\{f(t_i) < \theta - \gamma\} \leq \frac{2}{\ell}\left[(\left\lceil \mathrm{log}\mathcal{N}(\gamma, \mathcal{F}, 2\ell)\right\rceil+log\frac{2\ell}{\delta}\right],
\end{equation}
where $\mathcal{N}(\gamma, \mathcal{F}, 2\ell)=\text{sup}_{\mathcal{T}^{2\ell}} \mathcal{N}_{d}(\gamma, \mathcal{F})$, and $\mathcal{N}_d(\gamma, \mathcal{F})$ is the $\gamma-$covering number of function set $\mathcal{F}$ with the distance $d$ defined by 
\begin{equation}\label{dis}
d(f,g)=\mathrm{max}_{i\in [2\ell]}|f(t_i)-g(t_i)|, 
\end{equation}
where $f, g \in \mathcal{F}$. From Eq.~\ref{firstrisk} we can see that if the real images can be mapped to $\mathcal{S}$, $P\{f(t_i) < \theta - \gamma\}$ should be small. The number of real images $\ell$ in the sequence $\mathcal{T}$ and the choice of mapping function $f \in \mathcal{F}$, or more specifically $\Phi$ will affect the value of $P$. 

\par  We can also analyze the probability that a real image $I\notin \mathcal{T}$ is falsely mapped outside the constructed space $\mathcal{S}$. We fix $\theta \in \mathbb{R}$ and suppose that the range of the function set $\mathcal{F}$ is $[a,b]$. According to Theorem 9 in~\cite{scholkopf2001estimating}, with probability $1-\delta$ over the sequence $\mathcal{T}$, for all $\gamma > 0$ and any $f \in \mathcal{F}$,
\begin{equation}\label{secondrisk}
    P\{f(I)<\theta-\gamma \ \mathrm{and} \ I\notin \mathcal{T}\} \leq \frac{2}{\ell}\left(\Omega+\mathrm{log}\frac{4\ell}{\delta}\right),
\end{equation}
where 
\begin{equation}
\begin{split}
\Omega= \left \lceil \vphantom{\mathrm{log}\left (\frac{32\ell{(b-a)}^2}{\gamma^2} \right)} \mathrm{log} \mathcal{N} \left (\frac{\gamma}{2}, \mathcal{F}, 2\ell \right) \frac{64(b-a)\mathcal{D}(\mathcal{T},f,\gamma)}{\gamma^2} \right.\\ 
\left. \cdot \mathrm{log}\frac{e\ell\gamma}{8\mathcal{D}(\mathcal{T},f,\gamma)} \cdot \mathrm{log}\frac{32\ell{(b-a)}^2}{\gamma^2} \right \rceil,
\end{split}
\end{equation}
and $\mathcal{D}$ is defined as the sum of the distances from all real images $t_i \in \mathcal{T}$ falsely classified as generated images to the decision boundary: 
\begin{equation}
\mathcal{D}(\mathcal{T},f,\gamma)=\sum_{t_i \in \mathcal{T}} \mathrm{max}\{0,\theta+\gamma-f(t_i)\}.
\end{equation}
Eq.~\ref{secondrisk} shows that the probability $P\{f(I)<\theta-\gamma \ \mathrm{and} \ I\notin \mathcal{T}\}$ is bounded by the term of the ratio of the logarithmic covering numbers which is at scale proportional to $\gamma$, the size $\ell$ of the real image sequence $\mathcal{T}$. 

\par As shown in Fig.~\ref{mapping_function}(a1)$\sim$(a3), different values of $\ell$ (the number of real images in $\mathcal{T}$) will affect the decision boundary and consequently the number of false classifications. Also, the choice of mapping function $f$ decides the term $\mathcal{N}(\gamma, \mathcal{F}, 2\ell)$. Fig.~\ref{mapping_function}(b1)$\sim$(b3) show the classification results of using different $f$ while $\ell$ is fixed at 500. Therefore, for making the $\gamma-$covering $\mathcal{N}_d$ take a smaller value, $f$ should reflect the commonality of the real images so that it can map $\mathcal{T}$ to a dense subspace $\mathcal{S}$. 

\begin{figure}[t]
	\centering
	\includegraphics[height=10cm]{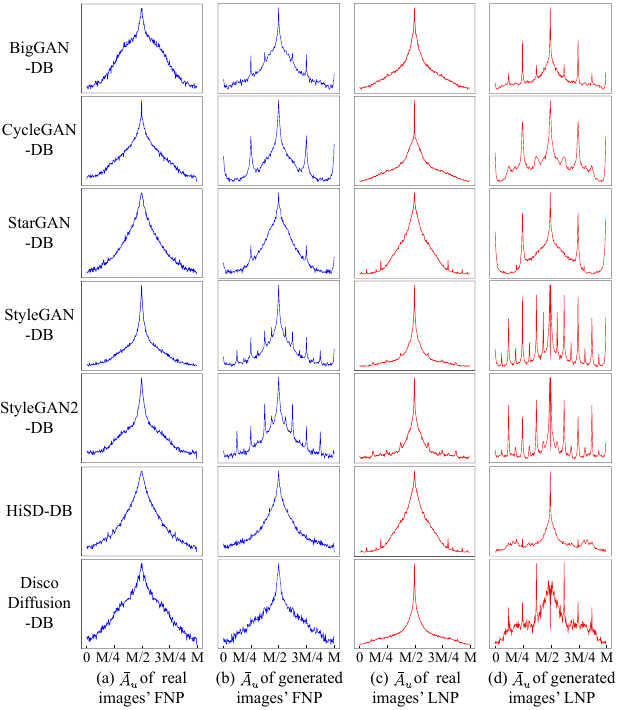}
 	\caption{The noise pattern (amplitude spectrum) of real images and generated images. The blue lines show the averaged FNP amplitude spectra $\bar{A}_u$. The red lines present the averaged LNP amplitude spectra (before feature enhancement).}
	\label{filter_and_lnp}
\end{figure}




\subsection{The Commonality of Real Images}
\par Previous research~\cite{frank2020leveraging,durall2020watch} found that the upsampling operation in conditional generative models changes the noise characteristics of input real images. Moreover, noise transformation after initial sampling is also in the image generation process of unconditional models. Therefore, we will explore the commonality of real images from the noise characteristics.

\par To extract image noise $n$ from the image $I$, applying a denoising filter $\Xi(\cdot)$ is the most common practice:
\begin{equation}
    n(x,y)= I(x,y)-\Xi(I(x,y)),
\end{equation}
where $(x,y)$ denotes coordinates. We adopted the denoising filters proposed in~\cite{yin2019side}. These filters keep image edges very effectively in practice, and obtain the noise relatively unaffected by image semantics. To better analyze the characteristics of the Filtered Noise Pattern (FNP) $n(x,y)$, we transform it by the Discrete Fourier Transform (DFT):
\begin{equation}
\zeta(u,v) = \frac{1}{{MN}}\sum\limits_{x = 0}^{M - 1} {} \sum\limits_{y = 0}^{N - 1} {n(x,y){e^{ - i2\pi ux/M}}{e^{ - i2\pi vy/N}}},
\end{equation}
where $M \times N$ represents the image size. We then calculate the amplitude spectrum $A(u,v)$ of $\zeta(u,v)$: 
\begin{align}\label{as}
    A(u,v) = \sqrt {{Rp}^2_{u,v} + {Ip}^2_{u,v}},
\end{align}
where ${Rp}_{u,v}$ and ${Ip}_{u,v}$ denote the real and imaginary parts of $\zeta(u,v)$, respectively. 

\par Since the Fourier frequency spectrum is symmetric, we can average the amplitude spectra $A(u,v)$ in either one dimension to show its characteristics. We took $A_u$ as an instance and calculated it by 
\begin{equation}\label{weight_avg}
   A_u = \sum\limits_{v=0}^{N-1}A(u,v).
\end{equation}
We then averaged $A_u$ across all images in each image subset (the image subsets are detailed in the experiment section) and denoted it as $\bar{A}_u$. It can be seen in Fig.~\ref{filter_and_lnp}(b) that the $\bar{A}_u$ of generated images' FNP show different distributions and periodic patterns in some subsets. However, the $\bar{A}_u$ for the real images' FNP do not show periodic patterns and have similar distributions, as shown in Fig.~\ref{filter_and_lnp}(a).

\begin{figure*}[htbp]
	\centering
	\includegraphics[height=4.2cm]{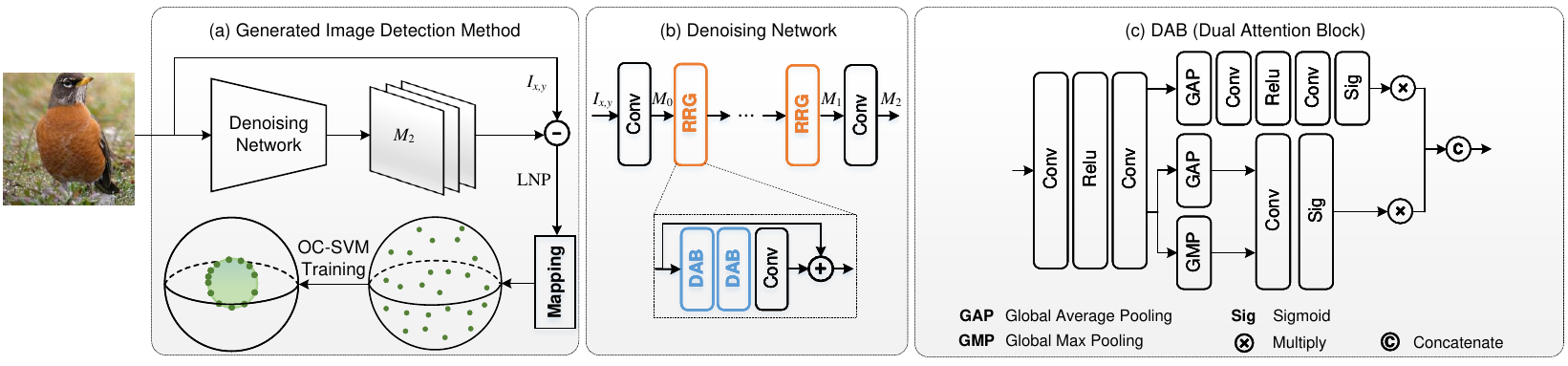}
	\caption{(a) Overall structure of the generated image detection method; (b) Structure of the denoising network for LNP extraction; (c) Structure of the Dual Attention Block (DAB), which is a component of the denoising network.}
	\label{fig:example3}
\end{figure*}

\par To confirm this commonality of real images, considering the strong learning ability of deep neural networks, we used a denoising network to extract image noise. For convenience, the equations only show the processing of one color channel, but we process all three channels of each color image similarly. The calculation can be formulated as 
\begin{equation} \label{lnpform}
    L(x,y) = I(x,y) - \Psi(I(x,y),{\theta_{\Psi}}), 
\end{equation} 
where $\Psi(\cdot)$ is a denoising network, $\theta_{\Psi}$ represents the set of all parameters of $\Psi(\cdot)$, and $\Psi(I(x,y))$ denotes the clean output image. $L(x,y)$ is called the Learned Noise Pattern (LNP). 

\par As shown in Fig.~\ref{filter_and_lnp}(c)(d), compared to FNP, LNP can better reflect the commonality of real images and their discrepancy from generated images. The $\bar{A}_u$ of generated images' LNP show periodic peaks whereas the $\bar{A}_u$ of real images' LNP do not. Moreover, the $\bar{A}_u$ of generated images' LNP from different models are distinct and show various characteristics. This explains why learning from one particular type of generated image causes a generalization problem. In contrast, the $\bar{A}_u$ of real images' LNP are much more similar. Therefore, mapping real images to the dense subspace constructed by noise features $f: \mathcal{T} \rightarrow \mathcal{S}$ is feasible and better reflects the commonality of real images. Consequently, the probability that a real image $t_i\in \mathcal{T}$ falls outside $\mathcal{S}:P\{f(t_i) < \theta - \gamma\}$ can be made small. 


\subsection{Mapping Real Images to a Dense Subspace}\label{3.1}

\par The general form of the LNP is described in Eq.~\ref{lnpform}, and there are many choices for the denoising network $\Psi( \cdot )$. Early denoising networks, such as DNCNN~\cite{zhang2017beyond}, directly add Gaussian white noise (AWGN) to clean images to create noisy images and then use the clean/noisy pairs to train $\Psi( \cdot )$. However, in real scenarios, the noise may be non-Gaussian. To extract the noise in the real world, CycleISP~\cite{zamir2020cycleisp} deploys networks to simulate the actual imaging pipeline of cameras by reconstructing RAW images from RGB images, adding noise to the RAW images, and then converting the RAW images to RGB images to get clean/noisy image pairs.

\begin{figure*}[htbp]
	\centering
	\includegraphics[height=8.5cm]{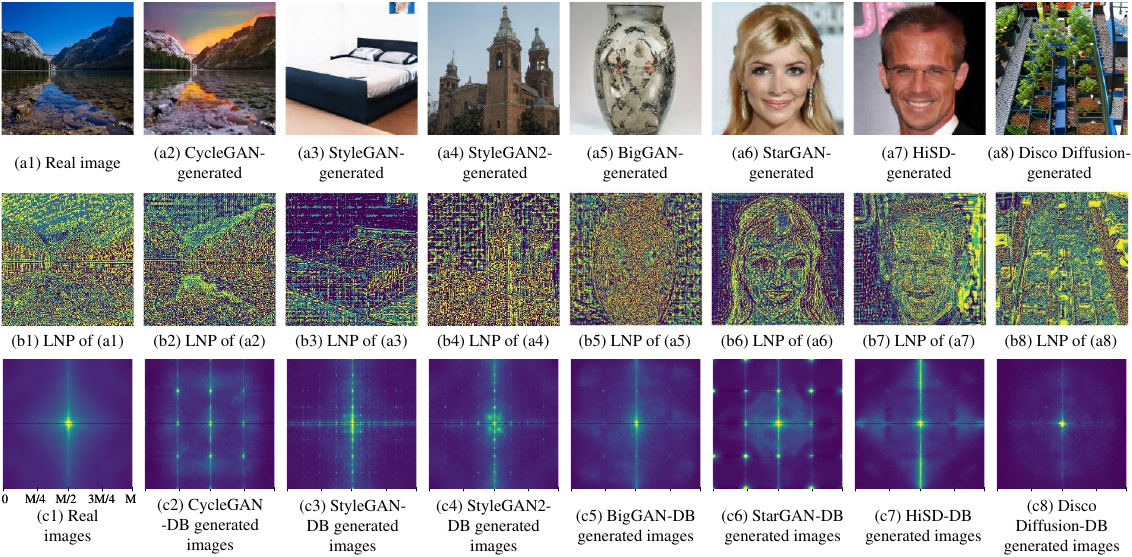}
	\caption{The discrepancy between real and generated images in the extracted LNP. (a) Real and generated images in datasets; (b) Visualization of LNPs of images in row a; (c) Average amplitude spectrum of real images from different real image subsets and that of generated images from different GAN models.}
	\label{fig:example4}
\end{figure*}

\par Inspired by the CycleISP~\cite{zamir2020cycleisp} network, the LNP extraction in our proposed generated image detection method is shown in Fig.~\ref{fig:example3}(b). Specifically, an image $I$ of size $M \times N$ will be processed as 
\begin{align} 
{M_0} = {K_3}(I(x,y)), 
\end{align} 
where ${K_3}$ denotes a $3 \times 3$ convolution. Many low-level applications including denoising~\cite{zhang2017beyond} benefit from a residual learning framework~\cite{he2016deep}, therefore we used the Recursive Residual Group (RRG) module to process the features. The RRG module is composed of two Dual Attention Blocks (DAB). Each DAB achieves calibration of the features by two types of channel attention and spatial attention, as shown in Fig.~\ref{fig:example3}(c). Its dual attention structure suppresses less useful features and keeps informative ones, which is ideal for denoising. We used four consecutive RRG modules:
\begin{align} 
{M_1} = RRG(RRG(RRG(RRG({M_0})))).
\end{align}
Finally, by a convolution operation ${M_2} = {K_3}({M_1})$, the three-channel feature map ${M_2}$ can be obtained, and then the LNP $L(x,y)$ in Eq.~\ref{lnpform} is calculated by 
\begin{align}    
L(x,y) = I(x,y) - {M_2}
 \label{NewLNP}
\end{align} 
in our generated image detection method.

\par Some LNP examples obtained by Eq.~\ref{NewLNP} are shown in Fig.~\ref{fig:example4}(b). We can see that the LNP in smooth regions of images from generative models exhibits grid artifacts, while the LNP of real images does not show that. Since a grid effect in the spatial domain will show periodicity in the frequency domain, we also transformed it by the DFT:
\begin{equation}
\zeta(u,v) = \frac{1}{{MN}}\sum\limits_{x = 0}^{M - 1} {} \sum\limits_{y = 0}^{N - 1} {L(x,y){e^{ - i2\pi ux/M}}{e^{ - i2\pi vy/N}}},
\end{equation}
and calculated amplitude spectra $A(u,v)$ by Eq.~\ref{as}. 

Similar to $\bar{A}_u$, we also averaged $A(u,v)$ of all images in each image subset and denote it as $\bar{A}(u,v)$. Fig.~\ref{fig:example4}(c) shows $\bar{A}(u,v)$ for each generated subset and real images from all real subsets. For the BigGAN~\cite{brock2018large} and HiSD~\cite{li2021image} models using nearest-neighbor interpolation for upsampling, the period in the horizontal direction of $\bar{A}(u,v)$ is $M$/4, and the period in the vertical direction of $\bar{A}(u,v)$ is $N$/4. For StyleGAN~\cite{karras2019style} and StyleGAN2~\cite{karras2020analyzing} using bilinear interpolation for upsampling, the period in the horizontal direction of $\bar{A}(u,v)$ is $M$/8, and the period in the vertical direction of $\bar{A}(u,v)$ is $N$/8. For CycleGAN~\cite{zhu2017unpaired} and StarGAN~\cite{choi2018stargan}, upsampling is performed using deconvolution, and the $\bar{A}(u,v)$ have strong vibrations with a period of $M$/4 in the horizontal direction of $\bar{A}(u,v)$ and a period of $N$/4 in the vertical direction. $\bar{A}(u,v)$ of all generated images' LNP show significant periodicity. In contrast, the $\bar{A}(u,v)$ of real images' LNP does not present periodicity. Therefore, we can calculate the amplitude spectrum $A(u,v)$ of LNP and then extract features from it to characterize the periodic signal.


\par To make the amplitude spectrum $A(u,v)$ of LNP more discriminative, we performed spectrum enhancement by increasing the gap between the high and low energy of the spectrum. Since the discriminative high peaks in Fig.~\ref{filter_and_lnp} are obtained by spectrum row averaging (Eq.~\ref{weight_avg}), amplitudes less than the average row value are not bright spots, and we set their value to 0. Then the differences of remaining non-zero amplitudes are enlarged by squaring. The enhancement is formulated as 
 \begin{align}
 {A'}(u,v)= \left\{ \begin{array}{l}
 0          \quad\ \ \ \ \ \ \ \ A(u,v)<A_u\\
 {A^2}(u,v) \quad otherwise
 \end{array} \right.,
 \label{enhancement}
 \end{align}
where $A_u$ is calculated by Eq.~\ref{weight_avg} on the amplitude spectrum $A(u,v)$ of LNP.

\par Since the LNP amplitude spectra of generated images present periodicity, and the LNP amplitude spectra of real images are generally low in energy except at the center, we extract features based on this property. The extracted features can be sampled by 
\begin{align} \label{SampF}   
F(m,n) = A'(u,v) \delta(u-{m  \Delta M},v-{n  \Delta N}),
\end{align} 
where $\delta(.)$ is a 2-D impulse function, $k$ is a sampling interval, and $m$ and $n$ are integers from $[0, \lfloor(M-1)/k\rfloor]$ and $[0,\lfloor(N-1)/k]\rfloor]$, respectively. The choice of $k$ will be discussed in the experiment. At this point, we can map real images $\mathcal{T}$ to a dense subspace $\mathcal{S}$ by $F(m,n)$ and finally optimize the one-class classifier.

\section{Experiments}\label{4}
\par We provide extensive experimental evidence to demonstrate the effectiveness of our method. This section introduces the datasets, describes the experimental setup and performance metrics, gives comparative experimental results, and presents a robustness study.

\begin{table*}[]
\renewcommand{\arraystretch}{1.6}
\renewcommand{\tabcolsep}{7.5pt}
\caption{Performance of a detection method that was sequentially fine-tuned by new model-generated images. Data in bold and underlined represents the best, and the evaluation metric is $ACC$.}

\begin{tabular}{l|ccccccccccc}
  \hline\hline
  \multirow{2}{*}{\diagbox[width=9em,trim=l]{Training}{Testing}} 
  & \multirow{2}{*}{\makecell[c]{Style\\GAN\\-DB}} 
  & \multirow{2}{*}{\makecell[c]{Style\\GAN2\\-DB}} 
  & \multirow{2}{*}{\makecell[c]{Cycle\\GAN\\-DB}} 
  & \multirow{2}{*}{\makecell[c]{Big\\GAN\\-DB}}  
  & \multirow{2}{*}{\makecell[c]{Star\\GAN\\-DB}} 
  & \multirow{2}{*}{\makecell[c]{HiSD\\-DB}} 
  & \multirow{2}{*}{\makecell[c]{Glow\\-DB}} 
  & \multirow{2}{*}{\makecell[c]{Latent\\Diffusion\\-DB}} 
  & \multirow{2}{*}{\makecell[c]{Disco\\Diffusion\\-DB}}  
  & \multirow{2}{*}{\makecell[c]{Stable\\Diffusion\\-DB}}  
  & \multicolumn{1}{c}{$ACC$} \\ \cline{12-12}
  &&&&&&&&&&& $avg.$ \\ \hline
StyleGAN-DB & \underline{\textbf{64.99}} & 60.35 & 50.08 & 51.40 & 49.97 & 49.92 & 49.97 & 50.25 & 50.00 & 49.86 & 52.68 \\
+ StyleGAN2-DB & 55.54 & \underline{\textbf{66.54}} & 50.29 & 50.45 & 50.00 & 50.03 & 49.97 & 50.10 & 52.00 & 54.22 & 52.91 \\
+ CycleGAN-DB  & 59.62 & 70.40 & \underline{\textbf{91.94}} & 51.62 & 51.95 & 60.03 & 47.87 & 50.60 & 47.00 & 40.87 & 57.19 \\
+ BigGAN-DB & 60.16 & 65.66 & 72.90 & \underline{\textbf{73.38}} & 44.72 & 49.40 & 44.32 & 52.60 & 56.00 & 44.14 & 56.33 \\
+ StarGAN-DB & 62.05 & 65.00 & 71.38 & 73.25 & \underline{\textbf{92.74}} & 50.63 & 49.52 & 53.05 & 54.00 & 48.77 & 62.04 \\
+ HiSD-DB & 60.62 & 61.64 & 67.67 & 68.08 & 90.29 & \underline{\textbf{53.13}} & 45.62 & 51.45 & 53.00 & 54.09 & 60.56 \\
+ Glow-DB & 61.16 & 62.05 & 68.85 & 67.41 & 92.39 & 52.28 & \underline{\textbf{88.75}} & 49.75 & 52.00 & 47.82 & 64.25 \\
+ Latent Diffusion-DB & 63.25 & 66.31 & 66.19 & 63.73 & 89.84 & 52.73 & 86.16 & \underline{\textbf{64.50}} & 54.00 & 51.09 & 65.78 \\ \hline\hline
\end{tabular}
\label{table:fig2}
\end{table*}


\subsection{Datasets and Experimental Setup}\label{4.1}

We trained and tested the proposed detection method on publicly available datasets whenever possible. We used five datasets provided by~\cite{wang2020cnn} containing real images and GAN-generated images from conditional GANs (BigGAN~\cite{brock2018large}, StarGAN~\cite{choi2018stargan}, CycleGAN~\cite{zhu2017unpaired}), and unconditional GANs (StyleGAN~\cite{karras2019style}, StyleGAN2~\cite{karras2020analyzing}) commonly used in generated image detection. To demonstrate the detection ability of our method on emerging generative models, we also included AE structure-based generative model HiSD~\cite{hisdgithub}, flow-based model Glow~\cite{kingma2018glow} whose network structure is reversible, Disco Diffusion \cite{discogithub} guided by CLIP~\cite{radford2021learning}, and the currently most advanced image generative diffusion models Stable Diffusion~\cite{rombach2022high} and Latent Diffusion \cite{latentgithub}. For the AE-based, Glow, and diffusion generative models, since public generated image datasets are not available, we used the officially published pre-trained models to produce generated image datasets without post-processing operations on the images. We denote the datasets using the model's name and a “-DB” suffix.


\par We randomly selected a certain number of real images from the real image subset provided by \cite{wang2020cnn}. The experiment shows that $\mathcal{T}$ with just 800 real images are enough to construct a dense subspace $\mathcal{S}$ by $F(m,n)$ and optimize the one-class classifier, which will be discussed in Subsection~\ref{sequence_t}. During training, the proposed generated image detection method did not see any generated images. All other generated image detection methods were implemented with their best performance. In the testing phase, we randomly selected 500 real and 500 generated images from each generative model dataset to infer their authenticity and evaluate our method and the other methods. All training was implemented on an NVIDIA GeForce GTX Titan X GPU and an Intel Xeon E5-2603 v4.

\subsection{Evaluation Metrics}\label{4.3}
\par Generated image detection can be seen as a binary classification problem. Each image belongs to one of two classes, real or generated, and a decision must be made for each image. Since our ultimate goal is to discern the authenticity of an image, Accuracy ($ACC$) is a proper evaluation metric for this task.

\par Precision in a binary classification task is related to the selection of classification thresholds. Therefore, we use the Average Precision ($AP$), which calculates the precision for all possible classification thresholds and then takes the average.

\par $F1$ score takes into account the Precision and Recall of the classification method. The $F1$ score can be considered a harmonic average of the Precision and Recall, with its maximum value of 1 and minimum value of 0.

\subsection{Verification of Fine-tuning with New Images}\label{futile}

\par Existing generated image detection methods may achieve satisfactory detection results in the laboratory. In practice, a detection method may face ever-growing generative models. Re-training a deployed detection network with existing and new model-generated images is possible, but the cost is too high in practice. A feasible solution is to use new model-generated images to fine-tune deployed detection networks. We experimented with this scenario to study whether this learning paradigm works. 

\par In our experiment, we used the generated image detection network proposed by~\cite{wang2020cnn}, which uses ResNet50 as the backbone. 5000 images from each generative model were used for fine-tuning, and 1000 per model were used for testing. The network was first trained with 5000 images from StyleGAN-DB. Then, images from datasets of StyleGAN2, CycleGAN, BigGAN, StarGAN, HiSD, Glow, and Latent Diffusion Models were used to fine-tune the network sequentially. After the initial training and each fine-tuning, we tested the network's detection performance on every generative model.  Results are in Table~\ref{table:fig2}.

\par We can see that as the new model-generated images are used to fine-tune the network, it tends to favor the newly seen generative model, while the detection accuracy of the previously seen generative models decreases, resulting in catastrophic forgetting~\cite{FRENCH1999128}. This experimental result is also consistent with our viewpoint depicted in Fig.~\ref{fig:ourmethod} that simply increasing the type of training set does not maintain or improve the test accuracy. Therefore, fine-tuning with new model-generated images does poorly and does not improve the model's generalization ability.

\subsection{Discussion of the Proposed Method}

\subsubsection{Effectiveness of LNP}

\begin{figure*}[t]
	\centering
	\includegraphics[height=8.6cm]{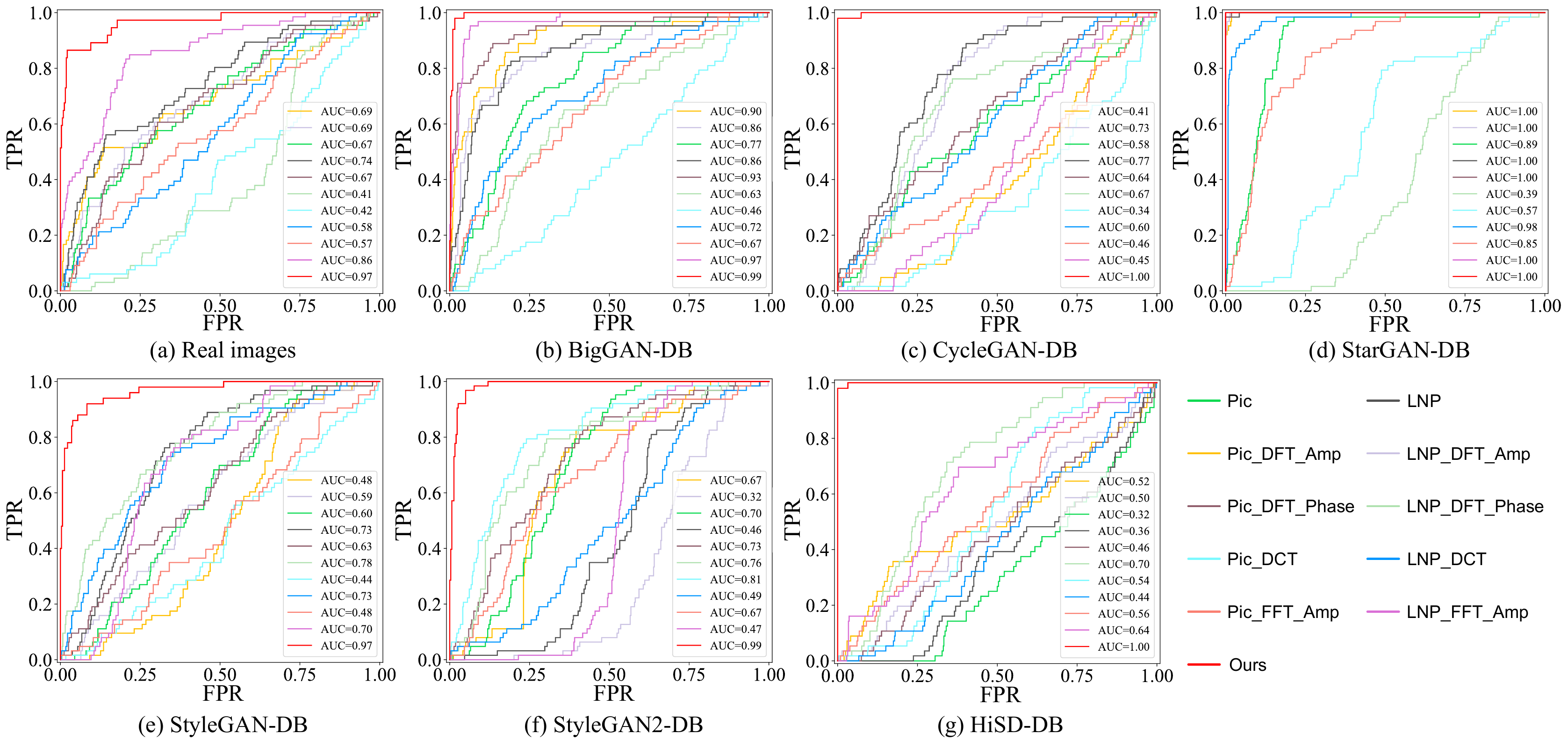}
	\caption{Comparison of discriminative ability of different features in a multi-class classification experiment. The results are shown as ROC curves. The seven classes are six generative models (StarGAN, BigGAN, CycleGAN, StyleGAN, StyleGAN2, HiSD) and real images.}
	\label{fig:example7}
\end{figure*}

\par Before mapping the real images $\mathcal{T}$ to a more dense subspace $\mathcal{S}$ by $F(m,n)$, we first verify the effectiveness of LNP. Using the enhanced LNP features to train common classifiers, such as SVM, can perform multi-class classification. In this way, we can verify the superiority of its discriminative ability over other image features. The experiment was set to perform seven-class classification on six generative models and real images. We randomly selected 500 generated images from each of the six generated image subsets and 500 real images from all six real image subsets, resulting in 3500 images in total.

\par We tested on many different types of image features: the original image (Pic), the amplitude spectrum after DFT of the original image (Pic\_DFT\_Amp), the phase spectrum after the DFT of the original image (Pic\_DFT\_Phase), the amplitude spectrum after the Fast Fourier Transform of the original image (Pic\_FFT\_Amp),  the original image after Discrete Cosine Transform (Pic\_DCT), the proposed LNP (LNP), the amplitude spectrum after the DFT of LNP (LNP\_DFT\_Amp), the phase spectrum after DFT of LNP (LNP\_DFT\_Phase), the amplitude spectrum after FFT of LNP (LNP\_FFT\_Amp), the LNP after DCT (LNP\_DCT), and the enhanced amplitude spectrum of LNP calculated by Eq.~\ref{enhancement} (Ours). The experimental result of five-fold cross-validation is shown by ROC curves in Fig.~\ref{fig:example7}. We see that using spatial features or phase spectra features cannot obtain stable performance in seven-class classification while using our proposed enhanced LNP feature extracted from amplitude spectra reaches over 0.97 in AUC and outperforms the other features mentioned above in detecting each generative model. The experiment shows that the proposed enhanced LNP features have stronger discriminative ability than the rest even in a more difficult classification task, laying a solid foundation for our generated image detection method.

\subsubsection{The Dimension of $F(m,n)$}\label{setk}
We sampled from the enhanced LNP amplitude spectra $A'(u,v)$ to construct $F(m.n)$. The value of $k$ affects the sampling frequency and consequently determines the dimensionality of extracted features $F(m,n)$. Excessive dimensionality brings redundant information, while too low dimensionality will lead to insufficient information, affecting the classification results. Therefore, we experimented to determine the value of $k$. The results are in Table~\ref{table:k}, and the choice of $k=32$ is reasonable. 

\begin{table}[t]
	\centering
	\renewcommand{\arraystretch}{1.5}
	\renewcommand{\tabcolsep}{5pt}
	\caption{Performance of our method under different dimension extracted feature $F(m,n)$ (determined by $k$). Data in bold and underlined represents the best.}
	\begin{tabular}{cccccccc}
		\hline \hline 
		Param & \multicolumn{6}{c}{Testing Set} & $ACC$ \\ \cmidrule(lr){1-1} \cmidrule(lr){2-7} \cmidrule(lr){8-8}
		\multirow{2}{*}{\emph{k}} & \multirow{2}{*}{\shortstack{Big\\GAN\\-DB}} & \multirow{2}{*}{\shortstack{Cycle\\GAN\\-DB}} & \multirow{2}{*}{\shortstack{HiSD\\-DB}} & \multirow{2}{*}{\shortstack{Star\\GAN\\-DB}} & \multirow{2}{*}{\shortstack{Style\\GAN\\-DB}} & \multirow{2}{*}{\shortstack{Style\\GAN2\\-DB}} &  \multirow{2}{*}{\emph{avg.}} 
            \\&  &  &  &  &  &  &  \\ 
		\hline
		4    & 0.585  & 0.820  & 0.845  & 0.845  & 0.760  & 0.890  & 0.791  \bigstrut[t]\\
		8    & 0.675  & 0.875  & 0.895  & 0.920  & 0.905  & 0.960  & 0.872  \\
		16   & 0.760  & 0.920  & 0.875  & 0.945  & 0.920  & 0.975  & 0.899  \\
		\rowcolor{gray!7} 32   & 0.810  & 0.980  & \underline{\textbf{0.900}} & 0.980  & \underline{\textbf{0.905}} & 0.980  & \underline{\textbf{0.926}} \\ 
		64   & \underline{\textbf{0.820}} & \underline{\textbf{0.990}} & 0.780  & \underline{\textbf{0.985}} & 0.715  & \underline{\textbf{0.990}} & 0.880  \bigstrut[b]\\
		\hline \hline 
	\end{tabular}%
	\label{table:k}%
 \vspace{1.5em}
\end{table}%

\begin{figure}[t]
	\centering
	\includegraphics[height=5.2cm]{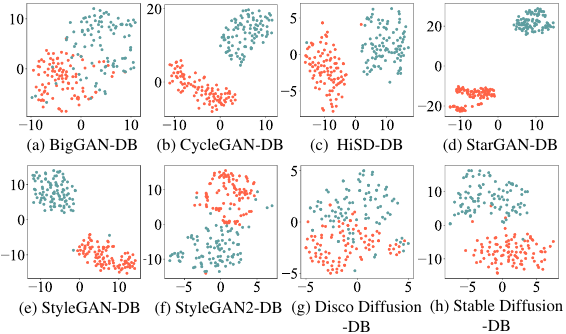}
	\caption{Visualization of $F(m,n)$ using t-SNE~\cite{van2008visualizing}. Red points denote real images, and green points denote generated images.}
	\label{fig:example6}
 \vspace{0.5em}
\end{figure}

For an image ($M=N=256$), when $k$ is 32, the final extracted feature $F(m,n)$ from the enhanced amplitude spectrum $A'(u,v)$ has only 64 dimensions (the maximum value of $m$ and $n$ is 7). To confirm this choice of $k$, we used t-SNE~\cite{van2008visualizing} to observe the feature distribution of the real images versus the generated images under eight datasets. We randomly selected 100 real images and 100 generated images from each dataset. The visualization results are in Fig.~\ref{fig:example6}. We see that the final sampled LNP feature $F(m,n)$  has a good discrimination ability. Moreover, it suggests that real images $\mathcal{T}$ can be mapped to a more dense subspace $\mathcal{S}$ by $F(m,n)$.

\subsubsection{Image Sequence $\mathcal{T}$}\label{sequence_t}
\par As discussed, we believe that real images have similarities, which enables one to construct a dense subspace $\mathcal{S}$ for one-class classification. To validate this view, we randomly selected one hundred to one thousand real images from the real image subsets in \cite{wang2020cnn} to construct $\mathcal{S}$. Then we tested it on every case and recorded the $ACC$ value. We experimented one hundred times in each case to prevent the coincidence of results. 

\par
\begin{figure}[t]
	\centering
	\includegraphics[height=3.7cm]{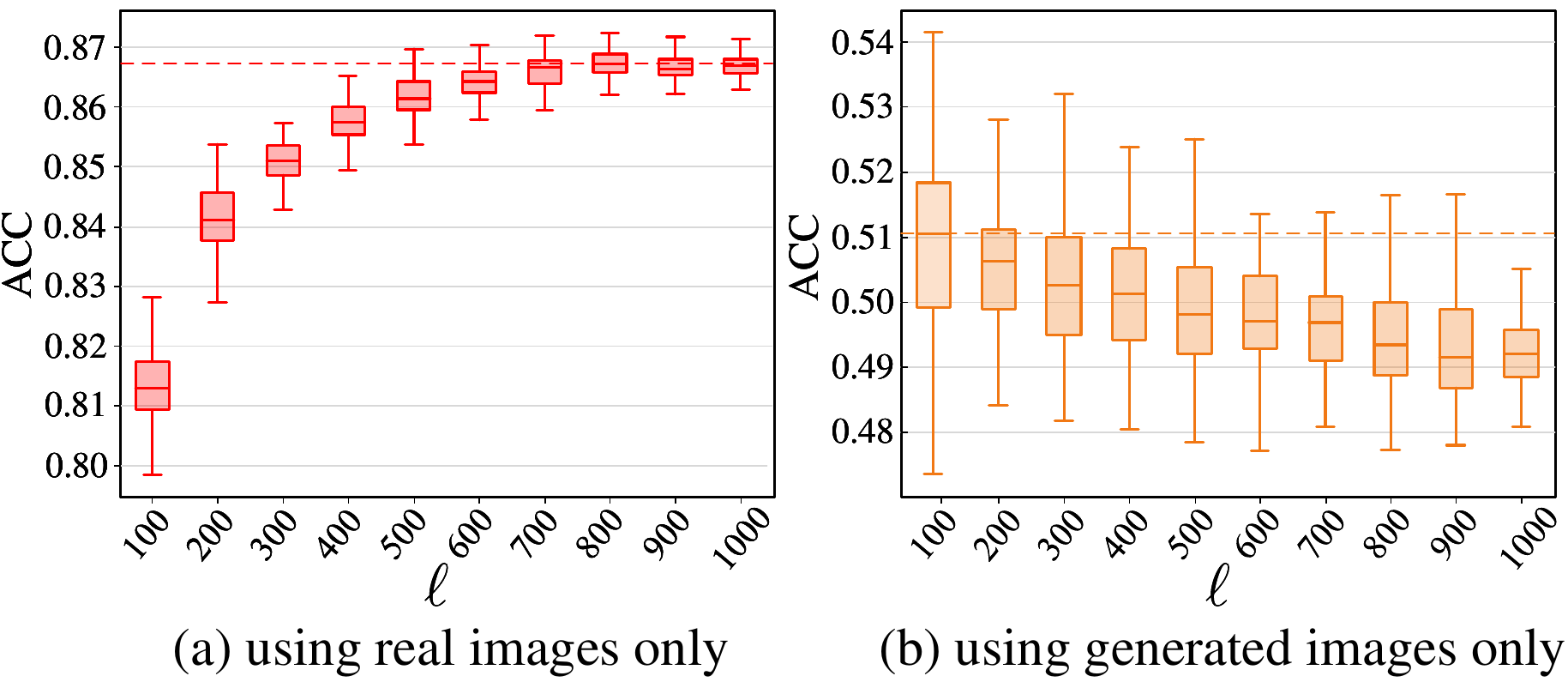}
	\caption{Types of images and their numbers $\ell$ in the image sequence $\mathcal{T}$ to construct a dense subspace $\mathcal{S}$ for one-class classification. (a) Only real images in $\mathcal{T}$. The dense subspace can be successfully constructed, and the best performance of one-class classification is achieved for $\ell = 800$. (b) Only generated images in $\mathcal{T}$. The constructed subspace is not sufficiently dense and one-class classification does not converge.}
	\label{fig:train_diff_num}
\end{figure}

 As shown in Fig.~\ref{fig:train_diff_num}(a), there is an optimal range of $\ell$ for mapping the real images $\mathcal{T}$ to a more dense subspace $\mathcal{S}$ by $F(m,n)$ discussed in Eq.~\ref{firstrisk} in Subsection~\ref{model_one_class}. Even when the number of real images in $\mathcal{T}$ is only 100, the $ACC$ of one-class classification reaches more than 80\%. As the number of real images $\ell$ increases, the results of one-class classification improve, and the best result is obtained when the number of real images $\ell$ is 800. On the other hand, since generated images are so different, they cannot be used to construct a subspace $\mathcal{S}$ for one-class classification, and the detection results are almost random guesses, as shown in Fig.~\ref{fig:train_diff_num}(b).

\par We also used real images from each generative model's real image subset as $\mathcal{T}$. 800 real images were randomly selected from each model's real image subset, and 100 real and 100 generated images were randomly selected from each model's real and generated image subsets to build the testing set. We repeated this process five times. The numbers in Table~\ref{table:Generalizability} are the average value of five experiments' results in the ACC metric.

\par Our method performs well in each case. We found that using the real images in the CycleGAN dataset as the image sequence $\mathcal{T}$ shows better performance. This is because the real images in the CycleGAN dataset include various categories, such as natural and biological images. 
In all our later experiments, we use 800 real images from the CycleGAN real image subset as the real images $\mathcal{T}$.

\begin{table}[t]
	\centering
	\renewcommand{\arraystretch}{1.5}
	\renewcommand{\tabcolsep}{3pt}
	\caption{Performance of our method when the image sequence $\mathcal{T}$ comes from different real image subsets. Data in bold and underlined represents the best.}
	\begin{tabular}{cccccccc}
		\hline\hline
		\multirow{3}{*}{Image Sequence $\mathcal{T}$} & \multicolumn{6}{c}{Testing Set} & \multirow{1}{*}{$ACC$} \\ \cmidrule(lr){2-7}\cmidrule(lr){8-8}
		& \makecell[c]{Big\\GAN\\-DB} & \makecell[c]{Cycle\\GAN\\-DB} & \makecell[c]{HiSD\\-DB} & \makecell[c]{Star\\GAN\\-DB} & \makecell[c]{Style\\GAN\\-DB} & \makecell[c]{Style\\GAN2\\-DB} & \emph{avg.}\\
		\hline
		BigGAN-DB & 0.822 & 0.908 & 0.770 & 0.966 & 0.848 & 0.794 & 0.851  \bigstrut[t]\\ \rowcolor{gray!7}
		CycleGAN-DB & \underline{\textbf{0.834}} & \underline{\textbf{0.944}} & 0.792 & \underline{\textbf{0.976}} & 0.902 & 0.844 & \underline{\textbf{0.882}} \\ 
		HiSD-DB & 0.720  & 0.750  & \underline{\textbf{0.902}} & 0.944 & 0.654 & 0.646 & 0.769 \\
		StarGAN-DB & 0.716 & 0.748 & 0.900  & 0.946  & 0.654  & 0.655  & 0.770 \\
		StyleGAN-DB & 0.812 & 0.934 & 0.624  & 0.946 & \underline{\textbf{0.964}} & 0.920  & 0.867   \\
		StyleGAN2-DB & 0.772 & 0.896 & 0.660 & 0.944 & 0.894  & \underline{\textbf{0.962}}  & 0.855                 \\ \hline\hline
		\label{table:Generalizability}
	\end{tabular}
\end{table}

\subsection{Quantitative Evaluation and Comparison}\label{4.5}
\subsubsection{Accuracy of Generated Image Detection}

\begin{table*}[t]
	\centering
	\renewcommand{\arraystretch}{1.2}
	\renewcommand{\tabcolsep}{1.8pt}
	\caption{Our method is compared with other state-of-the-art methods. Real\&Generated(ProGAN) indicates the number of real and generated images in the training set. Throughput indicates the number of images that can be authenticated per second. Data in bold and underlined represents the best, while data with parentheses represents the second best.}
\begin{tabular}{ccccccccccccccccc}
\toprule [1 pt]\toprule [1 pt]
\multirow{3}{*}{Methods} & \multicolumn{2}{c}{Training Set} & \multirow{3}{*}{Metrics} & \multicolumn{10}{c}{Testing Set} & \multicolumn{1}{c}{$ACC$} & Throughput \\ 
\cmidrule(lr){2-3} \cmidrule(lr){5-14} \cmidrule(lr){15-15} \cmidrule(lr){16-16} 
 & Real & \makecell[c]{Generated\\(ProGAN)} &  & \makecell[c]{Big\\GAN\\-DB} & \makecell[c]{Cycle\\GAN\\-DB} & \makecell[c]{Star\\GAN\\-DB} & \makecell[c]{Style\\GAN\\-DB} & \makecell[c]{Style\\GAN2\\-DB} & \makecell[c]{HiSD\\-DB} & \makecell[c]{Glow\\-DB} & \makecell[c]{Disco\\Diffusion\\-DB} & \makecell[c]{Latent\\Diffusion\\-DB} & \makecell[c]{Stable\\Diffusion\\-DB} & \makecell[c]{$\emph{avg.}$} & image/s \\ \toprule [1 pt]
\multirow{3}{*}{\makecell[c]{Zhang(Spec)\\WIFS'19\\\cite{zhang2019detecting}}} &
\multirow{3}{*}{\begin{minipage}[b]{0.07\columnwidth}\raisebox{-.5\height}{\includegraphics[width=\linewidth]{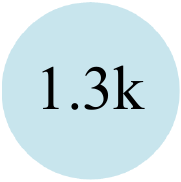}}
\end{minipage}} & 
\multirow{3}{*}{\begin{minipage}[b]{0.07\columnwidth}\raisebox{-.5\height}{\includegraphics[width=\linewidth]{Figures/13kcircles.pdf}}
\end{minipage}} & 
$ACC$ & 0.80 & 0.70 & \underline{\textbf{1.00}} & 0.62 & 0.55 & 0.56 & 0.65 & 0.37 & 0.65 & 0.62 & 0.65 & 
\multirow{3}{*}{\begin{minipage}[b]{0.055\columnwidth}\raisebox{-.5\height}{\includegraphics[width=\linewidth]{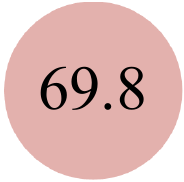}}
\end{minipage}} \\
 &  &  & $AP$ & 0.90 & 0.98 & \underline{\textbf{1.00}} & 0.68 & 0.59 & \underline{\textbf{1.00}} & (0.83) & 0.62 & 0.97 & 0.84 & 0.84 &  \\
 &  &  & $F1$ & 0.81 & 0.58 & \underline{\textbf{1.00}} & 0.64 & 0.60 & 0.70 & 0.48 & 0.54 & 0.74 & 0.72 & 0.68 &  \\
\toprule [0.3 pt]
\multirow{3}{*}{\makecell[c]{Zhang et al.\\WIFS'19\\\cite{zhang2019detecting}(Img)}} & 
\multirow{3}{*}{\begin{minipage}[b]{0.07\columnwidth}\raisebox{-.5\height}{\includegraphics[width=\linewidth]{Figures/13kcircles.pdf}}
\end{minipage}} & 
\multirow{3}{*}{\begin{minipage}[b]{0.07\columnwidth}\raisebox{-.5\height}{\includegraphics[width=\linewidth]{Figures/13kcircles.pdf}}
\end{minipage}} & 
$ACC$ & 0.55 & 0.91 & 0.83 & 0.58 & 0.68 & (0.96) & 0.50 & 0.49 & 0.53 & 0.49 & 0.65 & 
\multirow{3}{*}{\begin{minipage}[b]{0.055\columnwidth}\raisebox{-.5\height}{\includegraphics[width=\linewidth]{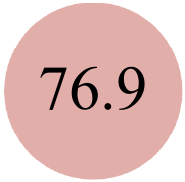}}
\end{minipage}} \\
 &  &  & $AP$ & 0.74 & 0.98 & (0.99) & 0.84 & 0.78 & \underline{\textbf{1.00}} & 0.53 & 0.39 & 0.66 & 0.42 & 0.73 &  \\
 &  &  & $F1$ & 0.67 & 0.90 & 0.86 & 0.70 & 0.70 & 0.96 & 0.45 & 0.65 & 0.67 & 0.65 & 0.72 &  \\
\toprule [0.3 pt]
\multirow{3}{*}{\makecell[c]{Frank et al.\\ICML'20\\\cite{frank2020leveraging}}} & 
\multirow{3}{*}{\begin{minipage}[b]{0.08\columnwidth}\raisebox{-.5\height}{\includegraphics[width=\linewidth]{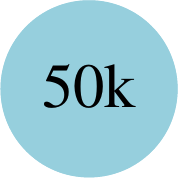}}
\end{minipage}} & 
\multirow{3}{*}{\begin{minipage}[b]{0.08\columnwidth}\raisebox{-.5\height}{\includegraphics[width=\linewidth]{Figures/50kcircles.pdf}}
\end{minipage}} & 
$ACC$ & 0.52 & 0.50 & 0.49 & 0.53 & 0.57 & 0.49 & 0.51 & 0.50 & 0.50 & 0.53 & 0.51 & 
\multirow{3}{*}{\begin{minipage}[b]{0.052\columnwidth}\raisebox{-.5\height}{\includegraphics[width=\linewidth]{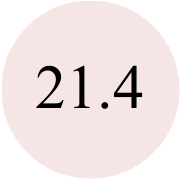}}
\end{minipage}} \\
 &  &  & $AP$ & 0.33 & 0.60 & 0.50 & 0.50 & 0.50 & 0.50 & 0.50 & 0.50 & 0.50 & 0.50 & 0.49 &  \\
 &  &  & $F1$ & 0.67 & 0.67 & 0.67 & 0.67 & 0.67 & 0.67 & 0.67 & 0.67 & 0.67 & 0.67 & 0.67 &  \\
\toprule [0.3 pt]
\multirow{3}{*}{\makecell[c]{Wang et al.\\CVPR'20\\\cite{wang2020cnn}}} & 

\multirow{3}{*}{\begin{minipage}[b]{0.09\columnwidth}\raisebox{-.5\height}{\includegraphics[width=\linewidth]{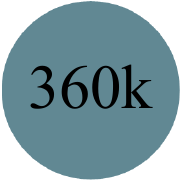}}
\end{minipage}} & 
\multirow{3}{*}{\begin{minipage}[b]{0.09\columnwidth}\raisebox{-.5\height}{\includegraphics[width=\linewidth]{Figures/360kcircles.pdf}}
\end{minipage}} & 
$ACC$ & 0.76 & 0.84 & 0.92 & 0.90 & 0.80 & 0.82 & 0.28 & 0.56 & 0.57 & 0.51 & 0.70 & 
\multirow{3}{*}{\begin{minipage}[b]{0.065\columnwidth}\raisebox{-.5\height}{\includegraphics[width=\linewidth]{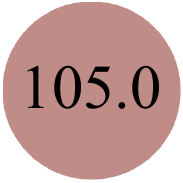}}
\end{minipage}} \\
 &  &  & $AP$ & (0.91) & 0.95 & (0.99) & (0.99) & \underline{\textbf{1.00}} & (0.99) & 0.35 & 0.55 & 0.94 & 0.83 & 0.85 &  \\
 &  &  & $F1$ & 0.80 & 0.88 & 0.95 & 0.85 & 0.93 & 0.94 & 0.57 & 0.67 & 0.71 & 0.67 & 0.80 &  \\
\toprule [0.3 pt]
\multirow{3}{*}{\makecell[c]{Chai et al.\\ECCV'20\\\cite{chai2020makes}}} & 
\multirow{3}{*}{\begin{minipage}[b]{0.09\columnwidth}\raisebox{-.5\height}{\includegraphics[width=\linewidth]{Figures/360kcircles.pdf}}
\end{minipage}} & 
\multirow{3}{*}{\begin{minipage}[b]{0.09\columnwidth}\raisebox{-.5\height}{\includegraphics[width=\linewidth]{Figures/360kcircles.pdf}}
\end{minipage}} & 
$ACC$ & 0.73 & 0.58 & 0.90 & 0.79 & 0.92 & 0.68 & 0.16 & 0.73 & 0.77 & 0.56 & 0.68 & 
\multirow{3}{*}{\begin{minipage}[b]{0.055\columnwidth}\raisebox{-.5\height}{\includegraphics[width=\linewidth]{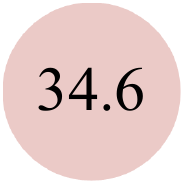}}
\end{minipage}} \\
 &  &  & $AP$ & 0.74 & 0.59 & \underline{\textbf{1.00}} & (0.99)& (0.99) & \underline{\textbf{1.00}} & 0.32 & 0.78 & 0.99 & (0.93) & 0.83 &  \\
 &  &  & $F1$ & 0.74 & 0.60 & 0.89 & 0.73 & 0.91 & 0.53 & 0.08 & 0.70 & 0.70 & 0.22 & 0.61 &  \\
\toprule [0.3 pt]

\multirow{3}{*}{\makecell[c]{Grag et al.\\ICME'21\\\cite{gragnaniello2021gan}}} & 
\multirow{3}{*}{\begin{minipage}[b]{0.09\columnwidth}\raisebox{-.5\height}{\includegraphics[width=\linewidth]{Figures/360kcircles.pdf}}
\end{minipage}} & 
\multirow{3}{*}{\begin{minipage}[b]{0.09\columnwidth}\raisebox{-.5\height}{\includegraphics[width=\linewidth]{Figures/360kcircles.pdf}}
\end{minipage}} & 
$ACC$ & 0.68 & 0.72 & \underline{\textbf{1.00}} & 0.89 & \underline{\textbf{0.99}} & \underline{\textbf{0.98}} & 0.43 & 0.66 & 0.69 & 0.54 & 0.76 & 
\multirow{3}{*}{\begin{minipage}[b]{0.048\columnwidth}\raisebox{-.5\height}{\includegraphics[width=\linewidth]{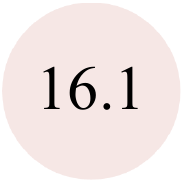}}
\end{minipage}} \\
 &  &  & $AP$ & 0.67 & 0.78 & \underline{\textbf{1.00}} & (0.99) & \underline{\textbf{1.00}} & \underline{\textbf{1.00}} & 0.35 & 0.72 & 0.80 & 0.76 & 0.81 &  \\ &  &  & $F1$ & 0.57 & 0.62 & \underline{\textbf{1.00}} & 0.90 & \underline{\textbf{0.99}} & \underline{\textbf{0.98}} & 0.16 & 0.69 & 0.72 & 0.71 & 0.73 &  \\
\toprule [0.3 pt]

\multirow{3}{*}{\makecell[c]{Liu et al.\\ECCV'22\\\cite{liu2022detecting}}} &
\multirow{3}{*}{\begin{minipage}[b]{0.09\columnwidth}\raisebox{-.5\height}{\includegraphics[width=\linewidth]{Figures/360kcircles.pdf}}
\end{minipage}} & 
\multirow{3}{*}{\begin{minipage}[b]{0.09\columnwidth}\raisebox{-.5\height}{\includegraphics[width=\linewidth]{Figures/360kcircles.pdf}}
\end{minipage}} & 
$ACC$ & \underline{\textbf{0.88}} & (0.92) & \underline{\textbf{1.00}} & \underline{\textbf{0.96}} &  0.92 & 0.94 & \underline{\textbf{0.80}} & 0.80 & 0.86 & 0.73& \underline{\textbf{0.88}} & 
\multirow{3}{*}{\begin{minipage}[b]{0.065\columnwidth}\raisebox{-.5\height}{\includegraphics[width=\linewidth]{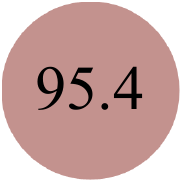}}
\end{minipage}} \\
 &  &  & $AP$ & \underline{\textbf{0.95}} & 0.98 & \underline{\textbf{1.00}} & (0.99) & (0.99) & \underline{\textbf{1.00}} & 0.69 & \underline{\textbf{0.97}} & 0.97 & \underline{\textbf{0.96}} & \underline{\textbf{0.95}} &  \\
 &  &  & $F1$ & \underline{\textbf{0.88}} & 0.91 & \underline{\textbf{1.00}} & \underline{\textbf{0.96}} & 0.93 & (0.96) & \underline{\textbf{0.80}} & (0.83) & 0.88 & 0.78& \underline{\textbf{0.89}} &  \\
\toprule [0.3 pt]

\multirow{3}{*}{\makecell[c]{Tan et al.\\CVPR'23\\\cite{tan2023learning}}} &
\multirow{3}{*}{\begin{minipage}[b]{0.08\columnwidth}\raisebox{-.5\height}{\includegraphics[width=\linewidth]{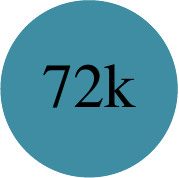}}
\end{minipage}} & 
\multirow{3}{*}{\begin{minipage}[b]{0.08\columnwidth}\raisebox{-.5\height}{\includegraphics[width=\linewidth]{Figures/72kcircles.pdf}}
\end{minipage}} & 
$ACC$ & 0.80 & 0.85 & (0.99) & (0.92) & (0.94) & 0.95 & 0.43 & \underline{\textbf{0.87}} & \underline{\textbf{0.98}} & (0.78)& (0.85) & 
\multirow{3}{*}{\begin{minipage}[b]{0.055\columnwidth}\raisebox{-.5\height}{\includegraphics[width=\linewidth]{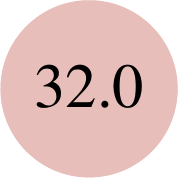}}
\end{minipage}} \\
 &  &  & $AP$ & 0.80 & 0.94 & \underline{\textbf{1.00}} & \underline{\textbf{1.00}} & \underline{\textbf{1.00}} & \underline{\textbf{1.00}} & 0.42 & (0.86) & \underline{\textbf{1.00}} & \underline{\textbf{0.96}} & (0.90) &  \\
 &  &  & $F1$ & 0.83 & 0.85 & (0.99) & (0.91) & (0.94) & 0.95 & 0.43 & \underline{\textbf{0.87}} & \underline{\textbf{0.98}} & 0.72& 0.85 &  \\
\toprule [0.3 pt]

\multirow{3}{*}{\makecell[c]{Ours\\-PNGAN \\ \cite{cai2021learning}}} & 
\multirow{3}{*}{\begin{minipage}[b]{0.065\columnwidth}\raisebox{-.5\height}{\includegraphics[width=\linewidth]{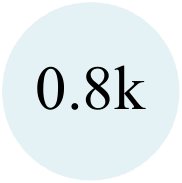}}
\end{minipage}} & 
& $ACC$ & 0.78 & \underline{\textbf{0.95}} & \underline{\textbf{1.00}} & \underline{\textbf{0.96}} & 0.51 & 0.60 & 0.52 & 0.54 & 0.81& 0.56 & 0.72 & 
\multirow{3}{*}{\begin{minipage}[b]{0.082\columnwidth}\raisebox{-.5\height}{\includegraphics[width=\linewidth]{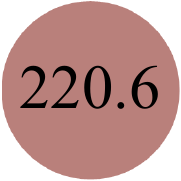}}
\end{minipage}} \\
 &  &  & $AP$ & 0.72 & \underline{\textbf{1.00}} & \underline{\textbf{1.00}} & \underline{\textbf{1.00}} & 0.59 & 0.79 & 0.50 & 0.59 & 0.91 & 0.55 & 0.77 &  \\
 &  &  & $F1$ & 0.80 & \underline{\textbf{0.95}} & \underline{\textbf{1.00}} & \underline{\textbf{0.96}} & 0.66 & 0.72 & 0.67 & 0.68 & 0.83 & 0.70 & 0.80 &  \\ \toprule [0.3 pt]

\multirow{3}{*}{\makecell[c]{Ours\\-Restormer\\ \cite{zamir2022restormer}}} & 
\multirow{3}{*}{\begin{minipage}[b]{0.065\columnwidth}\raisebox{-.5\height}{\includegraphics[width=\linewidth]{Figures/08kcircles.pdf}}
\end{minipage}} & 
& $ACC$ & 0.69 & 0.91 & 0.83 & (0.92) & 0.52 & 0.72 & (0.75) & 0.53 & 0.83 & \underline{\textbf{0.81}} & 0.75 & 
\multirow{3}{*}{\begin{minipage}[b]{0.087\columnwidth}\raisebox{-.5\height}{\includegraphics[width=\linewidth]{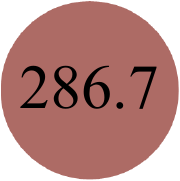}}
\end{minipage}} \\
 &  &  & $AP$ & 0.67 & (0.99) & \underline{\textbf{1.00}} & (0.99) & 0.62 & 0.80 & \underline{\textbf{0.84}} & 0.56 & 0.90 & 0.80 & 0.82 &  \\
 &  &  & $F1$ & 0.66 & 0.90 & 0.80 & (0.91) & 0.63 & 0.70 & (0.78) & 0.60 & 0.82 & \underline{\textbf{0.82}} & 0.76 &  \\ \toprule [0.3 pt]

\multirow{3}{*}{\makecell[c]{Ours\\-CycleISP\\ \cite{zamir2020cycleisp}}} & 
\multirow{3}{*}{\begin{minipage}[b]{0.065\columnwidth}\raisebox{-.5\height}{\includegraphics[width=\linewidth]{Figures/08kcircles.pdf}}
\end{minipage}} & 
& $ACC$ & (0.84) & \underline{\textbf{0.95}} & 0.98 & 0.90 & 0.85 & 0.79 & 0.66 & (0.81) & (0.93) & 0.74 & (0.85) & 
\multirow{3}{*}{\begin{minipage}[b]{0.10\columnwidth}\raisebox{-.5\height}{\includegraphics[width=\linewidth]{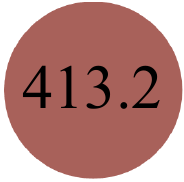}}
\end{minipage}} \\
 &  &  & $AP$ & 0.85 & \underline{\textbf{1.00}} & \underline{\textbf{1.00}} & \underline{\textbf{1.00}} & 0.98 & 0.84 & 0.74 & 0.78 & (0.97) & 0.67 & 0.88 &  \\
 &  &  & $F1$ & (0.85) & (0.94) & \ 0.98 & 0.89 & 0.93 & 0.83 & 0.76& 0.82 & (0.93) & (0.79) & {(0.87)} &  \\ \toprule [1 pt]\toprule [1 pt]

\end{tabular}
\label{table:Compare}
\end{table*}


\par We now use our proposed generated image detection method to authenticate images and compare it with recently proposed state-of-the-art methods. Our method is compatible with different denoising modules. We used two more denoising networks in the experiment: PNGAN~\cite{cai2021learning} and Restormer~\cite{zamir2022restormer}. Results are shown in Table~\ref{table:Compare}. Our method outperforms most methods in detection accuracy and is slightly behind our previous work~\cite{liu2022detecting}. However, it is worth noting that the proposed authentication method only takes 800 real images for training, using 0.1\% training images of that method, and avoids using any generated images while achieving good detection accuracy. We also note that Wang's approach~\cite{wang2020cnn} used 720k images as the training set and used ResNet50~\cite{he2016deep} as the classifier. The approach in~\cite{gragnaniello2021gan} used the same training set as~\cite{wang2020cnn} but used ResNet50~\cite{he2016deep} with two layers of downsampling removed as the classifier, and the training time is four times longer than~\cite{wang2020cnn}. In contrast, our method is of high efficiency in both the training and inference phase, able to cope with huge numbers of generated images with its 413.2 images/s throughput when using the CycleISP denoising module.

\par Moreover, our method can also effectively detect more types of generative models other than GAN models, such as Glow~\cite{kingma2018glow}, Disco Diffusion~\cite{discogithub}, Latent Diffusion~\cite{latentgithub}, and Stable Diffusion~\cite{rombach2022high}. Many existing generated image detection methods cannot effectively detect these model-generated images. In contrast, our method maintains good detection accuracy even though it did not see any images generated by them. This may help with the problem that detection algorithms always lag behind generative algorithms.

\begin{figure}[t]
	\centering
	\includegraphics[height=8.9cm]{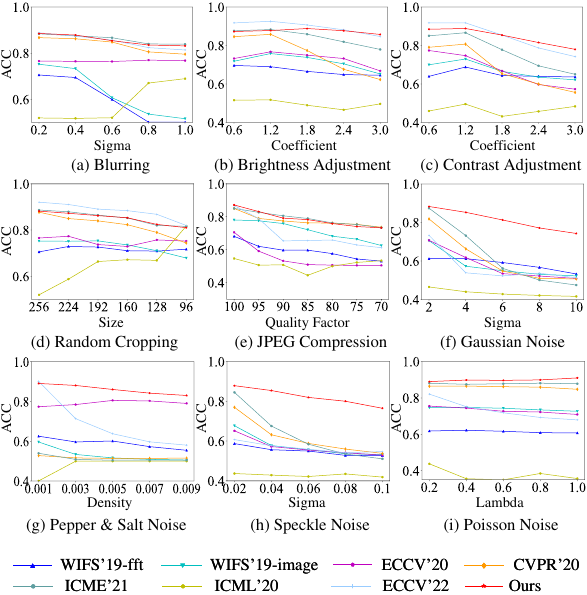}
	\caption{Robustness of our method against image blurring, brightness and contrast adjustment, random cropping, JPEG compression, corruption by different types of noise.}
	\label{fig:example9}
\end{figure}

\subsubsection{Robustness}

\par In real-world scenes, images are subjected to various post-processing operations, such as blurring and noise corruption. We tested the robustness of our method under different post-processing operations applied to generated images. The post-processing operations we used include:
\begin{itemize}
	\item Blurring: Gaussian filtering with a kernel size of 3 and sigma from 0.1 to 1.
        \item Brightness adjustment: the adjustment parameter is from 0.3 to 3. 
        \item Contrast adjustment: Gamma transform with $\gamma$ from 0.3 to 3.
        \item Random cropping: for a 256$\times$256 image, the cropping size was from 256 to 96, and we up-sampled the amplitude spectrum of the LNP of the final cropped image back to 256.
	\item JPEG compression: quality factors from 70 to 100.
	\item Gaussian noise: the sigma was set from 1 to 10, and the PSNR of the original image and the image after adding noise is from 26 to 47.
        \item Pepper \& Salt noise: the ratio of pepper to salt is 1:1. The density of the added noise is 0.001 to 0.01, and the PSNR of the noisy images is from 18 to 31.
        \item Speckle noise: the sigma ranges from 0.01 to 0.1, and the PSNR of the noisy images is from 22 to 57.
        \item Poisson noise: the lambda is set from 0.1 to 1, and the PSNR of the noisy images is from 3 to 58.
\end{itemize}

\par The one-class classifier we used was still constructed with 800 real images from CycleGAN-DB without any data enhancement via post-processing operations. Fig.~\ref{fig:example9} shows the robustness of our model via testing by post-processed images. We can see that our model has good robustness under various circumstances. The LNP amplitude spectrum generally is stable in terms of adding noise, blurring, and clipping. For JPEG compression, our method works stably with higher-quality compression factors. Since the DCT in JPEG operates on each 8$\times$8 patch, high compression leaves bright spots similar to generated images on the amplitude spectrum of images, resulting in a performance decline. 

\par
\begin{figure}[t]
	\centering
	\includegraphics[height=11cm]{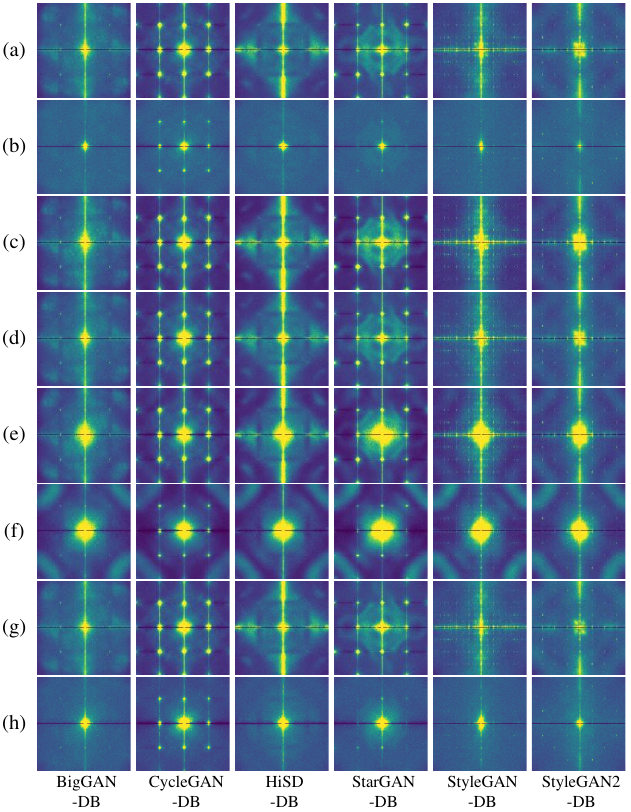}
	\caption{LNP's amplitude spectra of images with different levels of added noise. (a) Gaussian noise with $\sigma=1$; (b) Gaussian noise with $\sigma=10$; (c) Poisson noise with $\lambda=0.1$; (d) Poisson noise with $\lambda=1$; (e) Pepper and Salt noise with the density of 0.001; (f) Pepper \& Salt noise with density of 0.01; (g) Speckle noise with $\sigma=0.01$; (h) Speckle noise with $\sigma=0.1$.}
	\label{fig:example8}
\end{figure}

 In general, methods based on noise features are vulnerable to noise corruption. However, from the experimental results, our method shows robustness to different kinds of noise. The LNP is extracted by the denoising network and may contain added noise, as well as low-level features (grid artifacts) of generated images. However, the added noise is randomly distributed, so it does not mask the grid artifacts in the spatial domain, and the LNP remains highly periodic, which still differs from the characteristics of real images, as shown in Fig.~\ref{fig:example8}.

\begin{table}[t]
	\centering
	\renewcommand{\arraystretch}{1.5}
	\renewcommand{\tabcolsep}{2pt}
	\caption{The robustness of our method against spot suppression on amplitude spectra. We tested on real and generated images.}
	\begin{tabular}{cccccccccc}
		\hline\hline
		\multirow{2}{*}{scenario} & \multicolumn{3}{c}{All} & \multicolumn{3}{c}{Real} & \multicolumn{3}{c}{Generated}\\ \cmidrule(lr){2-4}\cmidrule(lr){5-7}\cmidrule(lr){8-10}
		& $ACC$ & $AP$ & $F1$ & $ACC$ & $AP$ & $F1$ & $ACC$ & $AP$ & $F1$ \\
		\hline
		unprocessed & 0.97 & 1.00 & 0.97 & 0.94 & 1.00 & 0.97 & 1.00 & 1.00 & 1.00  \\ 
		spot suppression & 0.97 & 1.00 & 0.97 & - & - & - & 1.00 & 1.00 & 1.00 \\ 
		\hline\hline
		\label{table:spotsuppression}
	\end{tabular}
\end{table}

\par We use the commonality of real images to detect generated images. However, future generative models will make generated images similar to real images as much as possible. Eliminating the bright spots in the amplitude spectrum can be used to reduce visual artifacts and deceive forensic algorithms. Durall's generative model~\cite{durall2020watch} based on DCGAN~\cite{radford2016unsupervised} uses a loss function to suppress spectrum bright spots during the training phase. We tested the robustness of our methods by discerning 100 real and 100 generated images randomly selected from their dataset~\cite{durall2020watch}. The experimental results shown in Table~\ref{table:spotsuppression} verify that spot suppression does not affect the detection accuracy.

\section{Conclusion}\label{6}

\par In this paper, we started from real images to detect generated images. We demonstrated that real images are very similar in their noise amplitude spectra. In contrast, generated images are very different. Therefore, generated images can be detected by mapping them out of the dense subspace constructed by real images only. The experimental results show that the method has good detection performance. The proposed method's superior detection performance and high efficiency allow its use in realistic scenarios, even for detecting possible future new models. Most importantly, this paper provides a new perspective on generated image detection.

\bibliography{bibfile}

\begin{thebibliography}{10}

\bibitem{liu2021psgan++}
Si~Liu, Wentao Jiang, Chen Gao, Ran He, Jiashi Feng, Bo~Li, and Shuicheng Yan.
\newblock Psgan++: Robust detail-preserving makeup transfer and removal.
\newblock {\em IEEE Transactions on Pattern Analysis and Machine Intelligence},
  44(11):8538--8551, 2021.

\bibitem{shen2020interfacegan}
Yujun Shen, Ceyuan Yang, Xiaoou Tang, and Bolei Zhou.
\newblock Interfacegan: Interpreting the disentangled face representation
  learned by gans.
\newblock {\em IEEE Transactions on Pattern Analysis and Machine Intelligence},
  44(4):2004--2018, 2020.

\bibitem{zhang2019detecting}
Xu~Zhang, Svebor Karaman, and Shih-Fu Chang.
\newblock Detecting and simulating artifacts in gan fake images.
\newblock In {\em Proceedings of the IEEE International Workshop on Information
  Forensics and Security (WIFS)}, pages 1--6. IEEE, 2019.

\bibitem{frank2020leveraging}
Joel Frank, Thorsten Eisenhofer, Lea Sch{\"o}nherr, Asja Fischer, Dorothea
  Kolossa, and Thorsten Holz.
\newblock Leveraging frequency analysis for deep fake image recognition.
\newblock In {\em Proceedings of the International Conference on Machine
  Learning}, pages 3247--3258. PMLR, 2020.

\bibitem{liu2020global}
Zhengzhe Liu, Xiaojuan Qi, and Philip~HS Torr.
\newblock Global texture enhancement for fake face detection in the wild.
\newblock In {\em Proceedings of the IEEE/CVF Conference on Computer Vision and
  Pattern Recognition}, pages 8060--8069, 2020.

\bibitem{dang2020detection}
Hao Dang, Feng Liu, Joel Stehouwer, Xiaoming Liu, and Anil~K Jain.
\newblock On the detection of digital face manipulation.
\newblock In {\em Proceedings of the IEEE/CVF Conference on Computer Vision and
  Pattern Recognition}, pages 5781--5790, 2020.

\bibitem{zhao2021multi}
Hanqing Zhao, Wenbo Zhou, Dongdong Chen, Tianyi Wei, Weiming Zhang, and Nenghai
  Yu.
\newblock Multi-attentional deepfake detection.
\newblock In {\em Proceedings of the IEEE/CVF Conference on Computer Vision and
  Pattern Recognition}, pages 2185--2194, 2021.

\bibitem{chai2020makes}
Lucy Chai, David Bau, Ser-Nam Lim, and Phillip Isola.
\newblock What makes fake images detectable? understanding properties that
  generalize.
\newblock In {\em Proceedings of the European Conference on Computer Vision},
  pages 103--120. Springer, 2020.

\bibitem{durall2020watch}
Ricard Durall, Margret Keuper, and Janis Keuper.
\newblock Watch your up-convolution: Cnn based generative deep neural networks
  are failing to reproduce spectral distributions.
\newblock In {\em Proceedings of the IEEE/CVF Conference on Computer Vision and
  Pattern Recognition}, pages 7890--7899, 2020.

\bibitem{wang2020cnn}
Sheng-Yu Wang, Oliver Wang, Richard Zhang, Andrew Owens, and Alexei~A Efros.
\newblock Cnn-generated images are surprisingly easy to spot... for now.
\newblock In {\em Proceedings of the IEEE/CVF Conference on Computer Vision and
  Pattern Recognition}, pages 8695--8704, 2020.

\bibitem{gragnaniello2021gan}
Diego Gragnaniello, Davide Cozzolino, Francesco Marra, Giovanni Poggi, and
  Luisa Verdoliva.
\newblock Are gan generated images easy to detect? a critical analysis of the
  state-of-the-art.
\newblock In {\em Proceedings of the IEEE International Conference on
  Multimedia and Expo (ICME)}, pages 1--6. IEEE, 2021.

\bibitem{yu2019attributing}
Ning Yu, Larry~S Davis, and Mario Fritz.
\newblock Attributing fake images to gans: Learning and analyzing gan
  fingerprints.
\newblock In {\em Proceedings of the IEEE/CVF International Conference on
  Computer Vision}, pages 7556--7566, 2019.

\bibitem{dzanic2020fourier}
Tarik Dzanic, Karan Shah, and Freddie Witherden.
\newblock Fourier spectrum discrepancies in deep network generated images.
\newblock {\em Advances in Neural Information Processing Systems},
  33:3022--3032, 2020.

\bibitem{karras2017progressive}
Tero Karras, Timo Aila, Samuli Laine, and Jaakko Lehtinen.
\newblock Progressive growing of gans for improved quality, stability, and
  variation.
\newblock {\em arXiv preprint arXiv:1710.10196}, 2017.

\bibitem{he2016deep}
Kaiming He, Xiangyu Zhang, Shaoqing Ren, and Jian Sun.
\newblock Deep residual learning for image recognition.
\newblock In {\em Proceedings of the IEEE/CVF Conference on Computer Vision and
  Pattern Recognition}, pages 770--778, 2016.

\bibitem{kingma2018glow}
Diederik~P Kingma and Prafulla Dhariwal.
\newblock Glow: Generative flow with invertible 1x1 convolutions.
\newblock {\em arXiv preprint arXiv:1807.03039}, 2018.

\bibitem{dinh2014nice}
Laurent Dinh, David Krueger, and Yoshua Bengio.
\newblock Nice: Non-linear independent components estimation.
\newblock {\em arXiv preprint arXiv:1410.8516}, 2014.

\bibitem{dinh2016density}
Laurent Dinh, Jascha Sohl-Dickstein, and Samy Bengio.
\newblock Density estimation using real nvp.
\newblock {\em arXiv preprint arXiv:1605.08803}, 2016.

\bibitem{papamakarios2017masked}
George Papamakarios, Theo Pavlakou, and Iain Murray.
\newblock Masked autoregressive flow for density estimation.
\newblock {\em Advances in Neural Information Processing Systems}, 30, 2017.

\bibitem{kingma2016improved}
Durk~P Kingma, Tim Salimans, Rafal Jozefowicz, Xi~Chen, Ilya Sutskever, and Max
  Welling.
\newblock Improved variational inference with inverse autoregressive flow.
\newblock {\em Advances in Neural Information Processing Systems}, 29, 2016.

\bibitem{sohl2015deep}
Jascha Sohl-Dickstein, Eric Weiss, Niru Maheswaranathan, and Surya Ganguli.
\newblock Deep unsupervised learning using nonequilibrium thermodynamics.
\newblock In {\em Proceedings of the International Conference on Machine
  Learning}, pages 2256--2265. PMLR, 2015.

\bibitem{song2019generative}
Yang Song and Stefano Ermon.
\newblock Generative modeling by estimating gradients of the data distribution.
\newblock {\em Advances in Neural Information Processing Systems}, 32, 2019.

\bibitem{ho2020denoising}
Jonathan Ho, Ajay Jain, and Pieter Abbeel.
\newblock Denoising diffusion probabilistic models.
\newblock {\em Advances in Neural Information Processing Systems},
  33:6840--6851, 2020.

\bibitem{rombach2021highresolution}
Robin Rombach, Andreas Blattmann, Dominik Lorenz, Patrick Esser, and Björn
  Ommer.
\newblock High-resolution image synthesis with latent diffusion models.
\newblock {\em arXiv preprint arXiv:2112.10752}, 2021.

\bibitem{rombach2022high}
Robin Rombach, Andreas Blattmann, Dominik Lorenz, Patrick Esser, and Bj{\"o}rn
  Ommer.
\newblock High-resolution image synthesis with latent diffusion models.
\newblock In {\em Proceedings of the IEEE/CVF Conference on Computer Vision and
  Pattern Recognition}, pages 10684--10695, 2022.

\bibitem{boser1992training}
Bernhard~E Boser, Isabelle~M Guyon, and Vladimir~N Vapnik.
\newblock A training algorithm for optimal margin classifiers.
\newblock In {\em Proceedings of the Fifth Annual Workshop on Computational
  Learning Theory}, pages 144--152, 1992.

\bibitem{scholkopf1999support}
Bernhard Sch{\"o}lkopf, Robert~C Williamson, Alex Smola, John Shawe-Taylor, and
  John Platt.
\newblock Support vector method for novelty detection.
\newblock {\em Advances in Neural Information Processing Systems}, 12, 1999.

\bibitem{tax2004support}
David~MJ Tax and Robert~PW Duin.
\newblock Support vector data description.
\newblock {\em Machine learning}, 54:45--66, 2004.

\bibitem{Zhai2023etd}
Wenbin Zhai, Liang Liu, Youwei Ding, Shanshan Sun, and Ying Gu.
\newblock Etd: An efficient time delay attack detection framework for uav
  networks.
\newblock {\em IEEE Transactions on Information Forensics and Security},
  18:2913--2928, 2023.

\bibitem{zhang2023federated}
Hengrun Zhang, Kai Zeng, and Shuai Lin.
\newblock Federated graph neural network for fast anomaly detection in
  controller area networks.
\newblock {\em IEEE Transactions on Information Forensics and Security},
  18:1566--1579, 2023.

\bibitem{lo2023adv}
Shao-Yuan Lo, Poojan Oza, and Vishal~M. Patel.
\newblock Adversarially robust one-class novelty detection.
\newblock {\em IEEE Transactions on Pattern Analysis and Machine Intelligence},
  45(4):4167--4179, 2023.

\bibitem{zaheer2022stab}
Muhammad~Zaigham Zaheer, Jin-Ha Lee, Arif Mahmood, Marcella Astrid, and
  Seung-Ik Lee.
\newblock Stabilizing adversarially learned one-class novelty detection using
  pseudo anomalies.
\newblock {\em IEEE Transactions on Image Processing}, 31:5963--5975, 2022.

\bibitem{scholkopf2001estimating}
Bernhard Sch{\"o}lkopf, John~C Platt, John Shawe-Taylor, Alex~J Smola, and
  Robert~C Williamson.
\newblock Estimating the support of a high-dimensional distribution.
\newblock {\em Neural Computation}, 13(7):1443--1471, 2001.

\bibitem{shawe1998structural}
John Shawe-Taylor, Peter~L Bartlett, Robert~C Williamson, and Martin Anthony.
\newblock Structural risk minimization over data-dependent hierarchies.
\newblock {\em IEEE Transactions on Information Theory}, 44(5):1926--1940,
  1998.

\bibitem{yin2019side}
Hui Yin, Yuanhao Gong, and Guoping Qiu.
\newblock Side window filtering.
\newblock In {\em Proceedings of the IEEE/CVF Conference on Computer Vision and
  Pattern Recognition}, pages 8758--8766, 2019.

\bibitem{zhang2017beyond}
Kai Zhang, Wangmeng Zuo, Yunjin Chen, Deyu Meng, and Lei Zhang.
\newblock Beyond a gaussian denoiser: Residual learning of deep cnn for image
  denoising.
\newblock {\em IEEE Transactions on Image Processing}, 26(7):3142--3155, 2017.

\bibitem{zamir2020cycleisp}
Syed~Waqas Zamir, Aditya Arora, Salman Khan, Munawar Hayat, Fahad~Shahbaz Khan,
  Ming-Hsuan Yang, and Ling Shao.
\newblock Cycleisp: Real image restoration via improved data synthesis.
\newblock In {\em Proceedings of the IEEE/CVF Conference on Computer Vision and
  Pattern Recognition}, pages 2696--2705, 2020.

\bibitem{brock2018large}
Andrew Brock, Jeff Donahue, and Karen Simonyan.
\newblock Large scale gan training for high fidelity natural image synthesis.
\newblock {\em arXiv preprint arXiv:1809.11096}, 2018.

\bibitem{li2021image}
Xinyang Li, Shengchuan Zhang, Jie Hu, Liujuan Cao, Xiaopeng Hong, Xudong Mao,
  Feiyue Huang, Yongjian Wu, and Rongrong Ji.
\newblock Image-to-image translation via hierarchical style disentanglement.
\newblock In {\em Proceedings of the IEEE/CVF Conference on Computer Vision and
  Pattern Recognition}, pages 8639--8648, 2021.

\bibitem{karras2019style}
Tero Karras, Samuli Laine, and Timo Aila.
\newblock A style-based generator architecture for generative adversarial
  networks.
\newblock In {\em Proceedings of the IEEE/CVF Conference on Computer Vision and
  Pattern Recognition}, pages 4401--4410, 2019.

\bibitem{karras2020analyzing}
Tero Karras, Samuli Laine, Miika Aittala, Janne Hellsten, Jaakko Lehtinen, and
  Timo Aila.
\newblock Analyzing and improving the image quality of stylegan.
\newblock In {\em Proceedings of the IEEE/CVF Conference on Computer Vision and
  Pattern Recognition}, pages 8110--8119, 2020.

\bibitem{zhu2017unpaired}
Jun-Yan Zhu, Taesung Park, Phillip Isola, and Alexei~A Efros.
\newblock Unpaired image-to-image translation using cycle-consistent
  adversarial networks.
\newblock In {\em Proceedings of the IEEE/CVF International Conference on
  Computer Vision}, pages 2223--2232, 2017.

\bibitem{choi2018stargan}
Yunjey Choi, Minje Choi, Munyoung Kim, Jung-Woo Ha, Sunghun Kim, and Jaegul
  Choo.
\newblock Stargan: Unified generative adversarial networks for multi-domain
  image-to-image translation.
\newblock In {\em Proceedings of the IEEE/CVF Conference on Computer Vision and
  Pattern Recognition}, pages 8789--8797, 2018.

\bibitem{hisdgithub}
Hisd.
\newblock \url{https://github.com/imlixinyang/HiSD}.
\newblock Accessed: 2023-07-12.

\bibitem{discogithub}
Disco diffusion.
\newblock \url{https://github.com/alembics/disco-diffusion}.
\newblock Accessed: 2023-07-12.

\bibitem{radford2021learning}
Alec Radford, Jong~Wook Kim, Chris Hallacy, Aditya Ramesh, Gabriel Goh,
  Sandhini Agarwal, Girish Sastry, Amanda Askell, Pamela Mishkin, Jack Clark,
  et~al.
\newblock Learning transferable visual models from natural language
  supervision.
\newblock In {\em Proceedings of the International Conference on Machine
  Learning}, pages 8748--8763. PMLR, 2021.

\bibitem{latentgithub}
Latent diffusion.
\newblock \url{https://github.com/CompVis/latent-diffusion}.
\newblock Accessed: 2023-07-12.

\bibitem{FRENCH1999128}
Robert~M. French.
\newblock Catastrophic forgetting in connectionist networks.
\newblock {\em Trends in Cognitive Sciences}, 3(4):128--135, 1999.

\bibitem{van2008visualizing}
Laurens Van~der Maaten and Geoffrey Hinton.
\newblock Visualizing data using t-sne.
\newblock {\em Journal of Machine Learning Research}, 9(11), 2008.

\bibitem{liu2022detecting}
Bo~Liu, Fan Yang, Xiuli Bi, Bin Xiao, Weisheng Li, and Xinbo Gao.
\newblock Detecting generated images by real images.
\newblock In {\em Proceedings of the European Conference on Computer Vision},
  pages 95--110. Springer, 2022.

\bibitem{tan2023learning}
Chuangchuang Tan, Yao Zhao, Shikui Wei, Guanghua Gu, and Yunchao Wei.
\newblock Learning on gradients: Generalized artifacts representation for
  gan-generated images detection.
\newblock In {\em Proceedings of the IEEE/CVF Conference on Computer Vision and
  Pattern Recognition}, pages 12105--12114, 2023.

\bibitem{cai2021learning}
Yuanhao Cai, Xiaowan Hu, Haoqian Wang, Yulun Zhang, Hanspeter Pfister, and
  Donglai Wei.
\newblock Learning to generate realistic noisy images via pixel-level
  noise-aware adversarial training.
\newblock {\em Advances in Neural Information Processing Systems},
  34:3259--3270, 2021.

\bibitem{zamir2022restormer}
Syed~Waqas Zamir, Aditya Arora, Salman Khan, Munawar Hayat, Fahad~Shahbaz Khan,
  and Ming-Hsuan Yang.
\newblock Restormer: Efficient transformer for high-resolution image
  restoration.
\newblock In {\em Proceedings of the IEEE/CVF Conference on Computer Vision and
  Pattern Recognition}, pages 5728--5739, 2022.

\bibitem{radford2016unsupervised}
Alec Radford, Luke Metz, and Soumith Chintala.
\newblock Unsupervised representation learning with deep convolutional
  generative adversarial networks.
\newblock {\em arXiv preprint arXiv:1511.06434}, 2016.

\end{thebibliography}

\end{document}